\documentclass[10pt,twocolumn,letterpaper]{article}
\usepackage[toc,page]{appendix}

\usepackage{wacv}              

\usepackage[accsupp]{axessibility}
\usepackage{graphicx}
\usepackage{amsmath}
\usepackage{amssymb}
\usepackage{booktabs}
\usepackage[normalem]{ulem}
\usepackage{multirow}
\usepackage[T1]{fontenc}
\usepackage{tabularx,ragged2e,booktabs}
\usepackage{tabularx} 
\usepackage{array} 
\newcolumntype{Y}{>{\centering\arraybackslash}X}
\newcolumntype{Z}{>{\centering\arraybackslash}m{0.93cm}}

\newcolumntype{M}{>{\centering\arraybackslash}m{1.52cm}}

\usepackage{colortbl}
\usepackage[dvipsnames]{xcolor}
\definecolor{Gray}{gray}{0.85}
\definecolor{SkyBlue}{rgb}{0.88,1,1}
\definecolor{PasteGreen}{RGB}{204, 226, 215}
\definecolor{PastePink}{RGB}{253, 223, 236}
\definecolor{PasteYellow}{RGB}{254, 255, 214}

\newcolumntype{T}{>{\columncolor{SkyBlue}}c}
\newcolumntype{D}{>{\columncolor{PasteYellow}}c}
\newcolumntype{R}{>{\columncolor{PastePink}}c}
\newcolumntype{E}{>{\columncolor{Gray}}c}

\usepackage[pagebackref,breaklinks,colorlinks]{hyperref}

\usepackage[capitalize]{cleveref}
\crefname{section}{Sec.}{Secs.}
\Crefname{section}{Section}{Sections}
\Crefname{table}{Table}{Tables}
\crefname{table}{Tab.}{Tabs.}

\usepackage{tikz}

\newcommand*\circled[1]{\tikz[baseline=(char.base)]{
            \node[shape=circle,draw,inner sep=0.25pt] (char) {#1};}}

\def\wacvPaperID{1026} 
\def\confName{WACV}
\def\confYear{2024}
\newcolumntype{C}{>{\Centering\hspace{0pt}}X}

\newlength\mylen
\newcolumntype{L}[1]{>{\RaggedRight\hspace{0pt}}p{#1}}

\begin{document}

\title{DashCop: Automated E-ticket Generation for Two-Wheeler Traffic Violations Using Dashcam Videos}

\author{
    Deepti Rawat\footnotemark[1], Keshav Gupta\footnotemark[1], Aryamaan Basu Roy, Ravi Kiran Sarvadevabhatla \\
    International Institute of Information Technology, Hyderabad, India \\
    {\tt\small deepti.rawat@research.iiit.ac.in, keshav.gupta@students.iiit.ac.in, basuroyaryamaan@gmail.com,} \\
     {\tt\small ravi.kiran@iiit.ac.in}
}

\maketitle
\footnotetext[1]{Authors contributed equally.}

\maketitle
\begin{abstract}
    Motorized two-wheelers are a prevalent and economical means of transportation, particularly in the Asia-Pacific region. However, hazardous driving practices such as triple riding and non-compliance with helmet regulations contribute significantly to accident rates. Addressing these violations through automated enforcement mechanisms can enhance traffic safety. In this paper, we propose DashCop, an end-to-end system for automated E-ticket generation. The system processes vehicle-mounted dashcam videos to detect two-wheeler traffic violations. Our contributions include: (1) a novel Segmentation and Cross-Association (SAC) module to accurately associate riders with their motorcycles, (2) a robust cross-association-based tracking algorithm optimized for the simultaneous presence of riders and motorcycles, and (3) the RideSafe-400 dataset, a comprehensive annotated dashcam video dataset for triple riding and helmet rule violations. Our system demonstrates significant improvements in violation detection, validated through extensive evaluations on the RideSafe-400 dataset. Project page: \url{https://dash-cop.github.io/}
\end{abstract}

\section{Introduction}
\label{sec:intro}

Motorized two-wheelers have been widely used as an affordable and effective commuting choice, especially in the Asia-Pacific region ~\cite{mordor2023,fortune2023,gmi2023,markntel2023}. However, hazardous driving habits, such as not wearing helmets, exceeding passenger limits, and wrong-side driving, often pose significant safety risks. Annually, more than half a million two-wheeler accidents are reported in this region~\cite{miah2024analyzing}. This alarming statistic underscores the need for promoting helmet compliance and deterring triple riding by imposing penalties. Such measures aim to curb dangerous driving practices, ensure rider safety, and maximize overall traffic safety goals~\cite{sharma2024two,barreto2021object}.

To address these safety concerns, there is a critical need for efficient enforcement mechanisms. Automated electronic traffic ticket (E-ticket) systems offer a promising solution by automating the detection of specific violations and using automatic number plate recognition (ANPR) to generate E-tickets~\cite{barreto2021object,thombre2022automated,daniel2024}. This automation reduces reliance on manual enforcement, which is particularly beneficial in countries with a low ratio of traffic police to vehicles~\cite{ghosh2023deep,chandravanshi2021automated,toi2019}.

Detecting the violation of interest is an essential component of these E-ticket generation systems. Many works have individually  addressed automating the detection of triple riding and helmet rule violations~\cite{duong2023helmet,tsai2023video,wang2023prb,aboah2023real,cui2023effective,tran2023robust,goyal2022detecting,soltanikazemi2023real}. These works process video frames from statically mounted cameras (\eg CCTV). Despite the advantage of continuous monitoring, their scope is limited in terms of coverage over the road network and high installation costs. Moreover, the fixed and predictable positions of the cameras lack the element of unpredictability that is crucial for effective deterrence.

\begin{figure*}[!t]
\begin{center}
\includegraphics[width=0.9\linewidth]{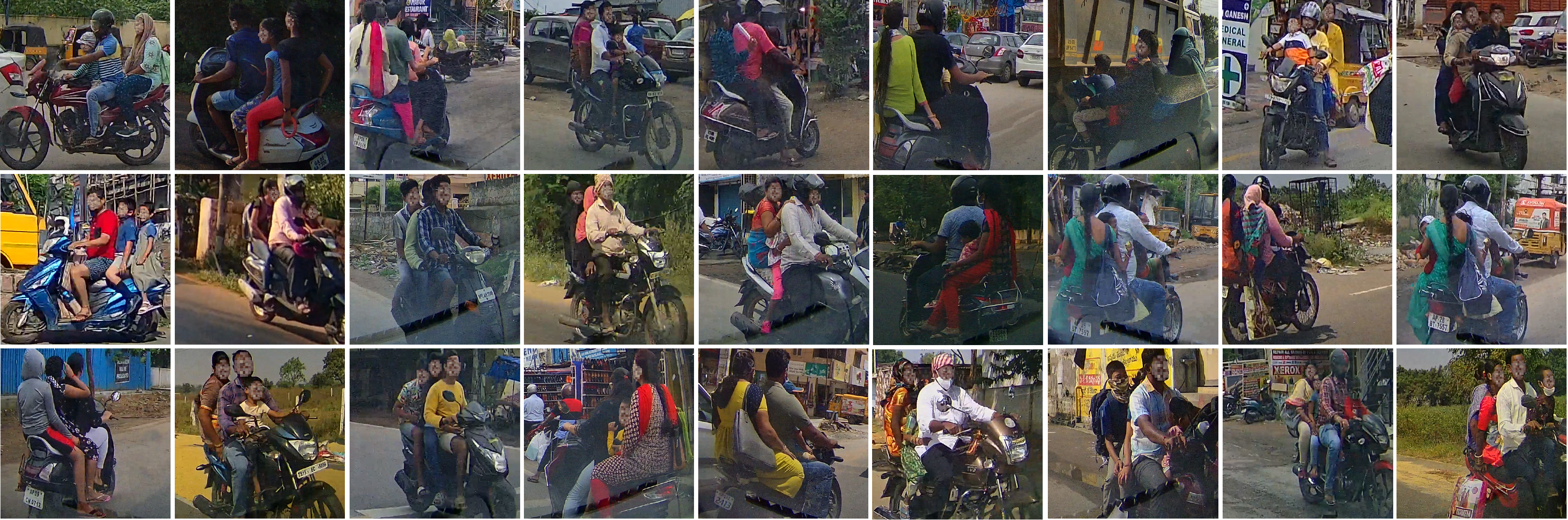}
   \caption{Instances of triple riding and helmet rule violations from our RideSafe-400 dataset.}
\label{fig:diverse_scenarios}
\end{center}
\end{figure*}

Vehicle-mounted dashcams offer a complementary solution to this challenge. They provide a broader range of road network coverage for traffic enforcement. Unlike static cameras, they have the advantage of unpredictability which is essential for effective deterrence. With the growing ubiquity of vehicle-mounted dashcams~\cite{yahoo2024,fmi2022,straits2022}, it is essential to develop systems that can effectively utilize dashcam videos for automated detection of traffic violations. 

Detecting two-wheeler traffic violations via dashcam footage presents numerous challenges. These include relative motion between the dashcam-equipped vehicle and other vehicles, the unpredictable dynamics of dense and unstructured traffic environments, occlusions caused by vehicles, riders, other road elements, and the small spatial footprint of objects of interest (\eg motorcycles, riders, license plates). The current body of work addressing dashcam-based detection of two-wheeler traffic violations is extremely limited~\cite{goyal2022detecting,peng2022traffic}, and no solution extends to the automated ticket generation stage. 

To bridge this gap, we propose an end-to-end system for automating the generation of E-tickets for two-wheeler traffic violations, utilizing dashcam videos as input. Our contributions include:
\begin{itemize}
\item \textbf{Segmentation and Cross-Association (SAC):} A novel segmentation module for associating riders with their corresponding two-wheeler.
\item \textbf{Association-based Tracker}: A novel instance tracking module with a formulation optimized for joint tracking of riders and associated two-wheelers.
\item \textbf{RideSafe-400 Dataset:} An annotated dashcam video dataset containing triple riding and helmet rule violations. The dataset includes extensive track-level mask and bounding box annotations for six object classes: \texttt{rider-motorcycle}, \texttt{rider}, \texttt{motorcycle}, \texttt{helmet}, \texttt{no-helmet}, \texttt{license plate}.
\end{itemize}

We use our dataset to evaluate the E-ticket generation system and to benchmark its various components. Please refer to the \url{https://dash-cop.github.io/} for videos and additional media related to our work.

\section{Related Works}
\label{sec:related}

\textit{Helmet Detection:} Previous research on helmet detection predominantly utilizes static cameras~\cite{vo2024robust,chen2024effective,van2024motorcyclist,zhang2024coarse,bouhayane2023swin,cui2023effective}. While some methods process the entire frame~\cite{vo2024robust,bouhayane2023swin,chen2024effective,van2024motorcyclist,duong2023helmet}, others utilize cropped images of the riding instance as input~\cite{zhang2024coarse,sugiarto2021helmet}. State-of-the-art detection frameworks such as YOLO~\cite{Jocher_Ultralytics_YOLO_2023} and DETR~\cite{carion2020end,zong2023detrs} are commonly employed. In our approach, we perform detection using the full frame and leverage YOLO-v8x~\cite{Jocher_Ultralytics_YOLO_2023} as the base detector.

\textit{Triple Riding Detection:} Similar to helmet detection, triple riding detection is typically performed using static cameras and established detection architectures~\cite{kumar2023triple,mallela2021detection,kumari2024effcient,joy2023traffic,choudhari2023traffic,nandhakumar2023yolo,chaturvedi2023detection}. Determining triple riders is done either by counting the number of riders belonging to a driving instance~\cite{goyal2022detecting,kumar2023triple,mallela2021detection,choudhari2023traffic,chaturvedi2023detection} or holistically, as a detection problem~\cite{charran2022two,nandhakumar2023yolo}. Our approach treats the task as a classification problem.

\textit{Rider-Motorcycle Association:} Determining the association between riders and their motorcycle is crucial for multiple downstream modules in the context of the violation detection task. Current methods often rely on the overlap between rider and motorcycle bounding boxes for rider-motorcycle (R-M) association~\cite{chaturvedi2023detection,choudhari2023traffic,mallela2021detection}. Goyal \etal~\cite{goyal2022detecting} introduce a learnable trapezium-based representation with IoU-threshold-based suppression for compact modeling. However, these methods treat rider and motorcycle detections independently, which can degrade association performance in dense traffic scenarios. Our approach explicitly models the spatial relationship between a motorcycle and its riders, resulting in more robust associations.

\textit{Tracking:} In most works, trackers based on the SORT~\cite{bewley2016simple} family such as DeepSORT~\cite{wojke2017simple} are employed~\cite{goyal2022detecting,van2024motorcyclist,duong2023helmet,choudhari2023traffic}. However, tracking is typically done at the individual rider or motorcycle level. We introduce a unique extension of the basic SORT tracker to enable collective tracking of riders and their corresponding motorcycles as a single associated entity.

\textit{End-to-end systems:} Dashcam footage has been used in various contexts~\cite{shaikh2024evaluating,rocky2024review,ariffin2021exploring,zekany2019classifying,kar2017real}, but its application for traffic violation detection is limited~\cite{goyal2022detecting,peng2022traffic,sugiarto2021helmet}. Automatic or semi-automatic systems for end-to-end E-ticket generation are usually based on static cameras~\cite{charran2022two,mallela2021detection}. However, these approaches often rely on image processing heuristics and are prone to fixed viewpoint bias. Also, the datasets tend to be fairly small and end-to-end evaluation (\ie E-ticket performance metrics) is absent. To the best of our knowledge, our work is the first to  introduce an end-to-end E-ticket generation system for two-wheelers \textit{using dashcam videos}.

\section{RideSafe-400 Dataset}
\label{sec:dataset}

Our dataset, RideSafe-400, comprises 400 driving videos specifically curated for identifying triple riding and helmet rule violators on the road. The traffic scenarios are captured using a DDPAI X2S Pro dashcam, with a resolution of $1920 \times 1080$ pixels and a frame rate of 25 fps. Each video spans between 60 to 72 seconds, culminating in approximately 600K frames. These videos contain multiple instances of triple riding or helmet rule violations, captured in a diverse range of traffic conditions, including varying traffic types, street types, illumination, weather, and other challenging scenarios (see \cref{fig:diverse_scenarios}).

\begin{figure}[!t]
  \centering  
   \includegraphics[width=0.9\linewidth]{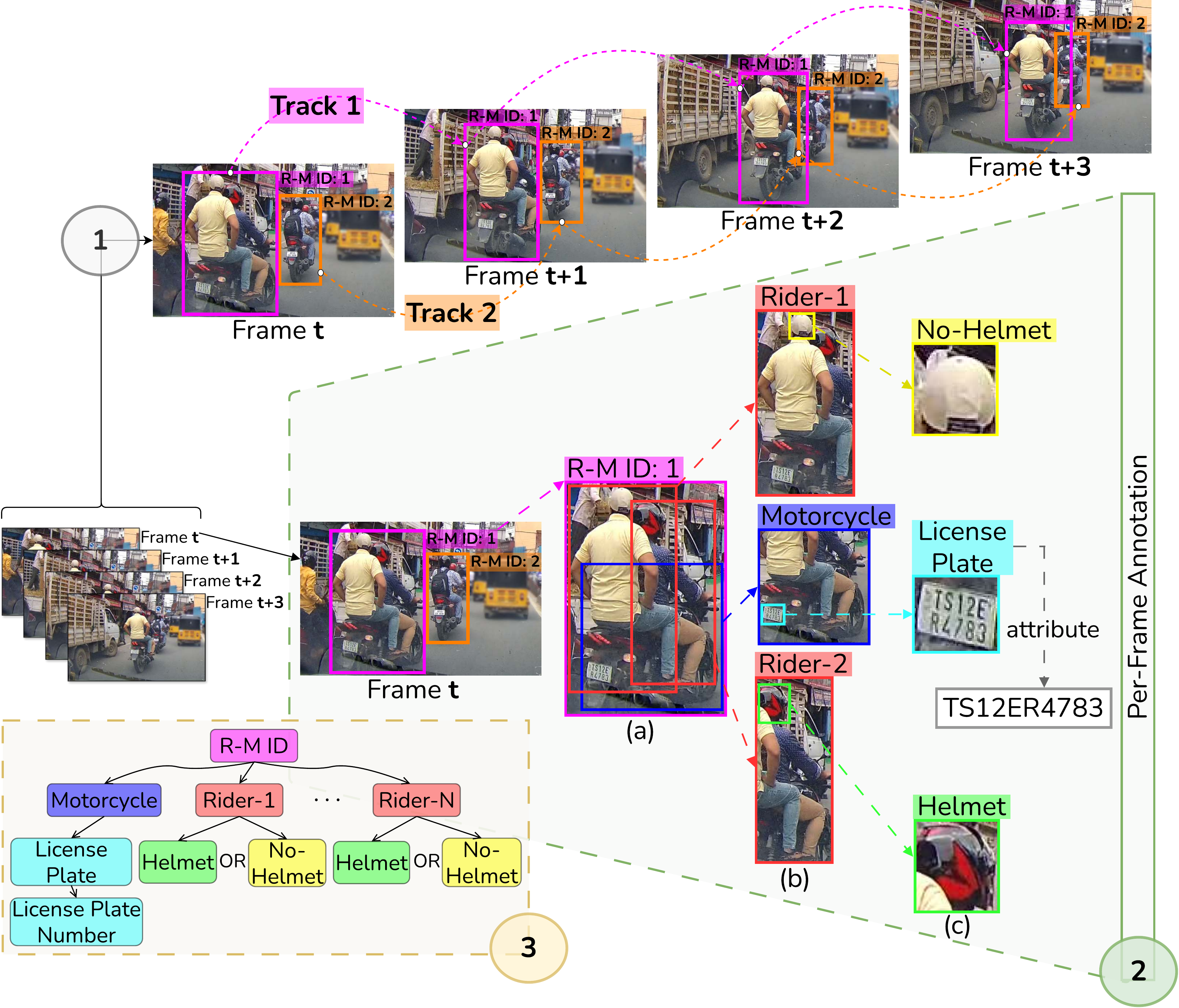}
   \caption{Annotation schema for our RideSafe-400 dataset. Refer to \cref{sec:dataset} for details.}
   \label{fig:annotation_classes}
\end{figure}

Annotations in RideSafe-400 cover six object classes: `R-M instance', `rider', `motorcycle', `helmet', `no-helmet', and `license plate' (see \cref{fig:annotation_classes}-\circled{1}). For each video frame, riders are associated with their respective motorcycles (termed an R-M instance) using rectangular bounding boxes. Each R-M instance is further annotated with the classes rider and motorcycle (see \cref{fig:annotation_classes}-\circled{2} (a)). For each rider, the presence or absence of a helmet is annotated (\cref{fig:annotation_classes}-\circled{2} (b)). Each motorcycle object is assigned an attribute `license plate' or `no license plate', depending on the visibility of the license plate, and the corresponding license plate bounding box is annotated along with its plate number (\cref{fig:annotation_classes}-\circled{2} (c)). To uniquely identify an R-M instance across video frames, a \textit{track id} is provided (see \cref{fig:annotation_classes}-\circled{1}). Additionally, each R-M instance and its corresponding objects (rider, motorcycle, helmet, no-helmet, license plate) are assigned a common \textit{group id} attribute (see \cref{fig:annotation_classes}-\circled{3}).

Our dataset includes 354K R-M annotations and 23K R-M tracks, 356K motorcycle annotations and 24K motorcycle tracks, 311K rider annotations and 19K rider tracks, 194K helmet annotations and 10K helmet tracks, 149K no-helmet annotations and 8K no-helmet tracks. Additionally, there are 27K annotated license plates, each labelled with its corresponding license plate number.
A detailed description of annotation procedure is included in Supplementary.

\section{Methodology}
\label{sec:methodology}

\begin{figure*}[!ht]
\begin{center}
\includegraphics[width=0.9\linewidth]{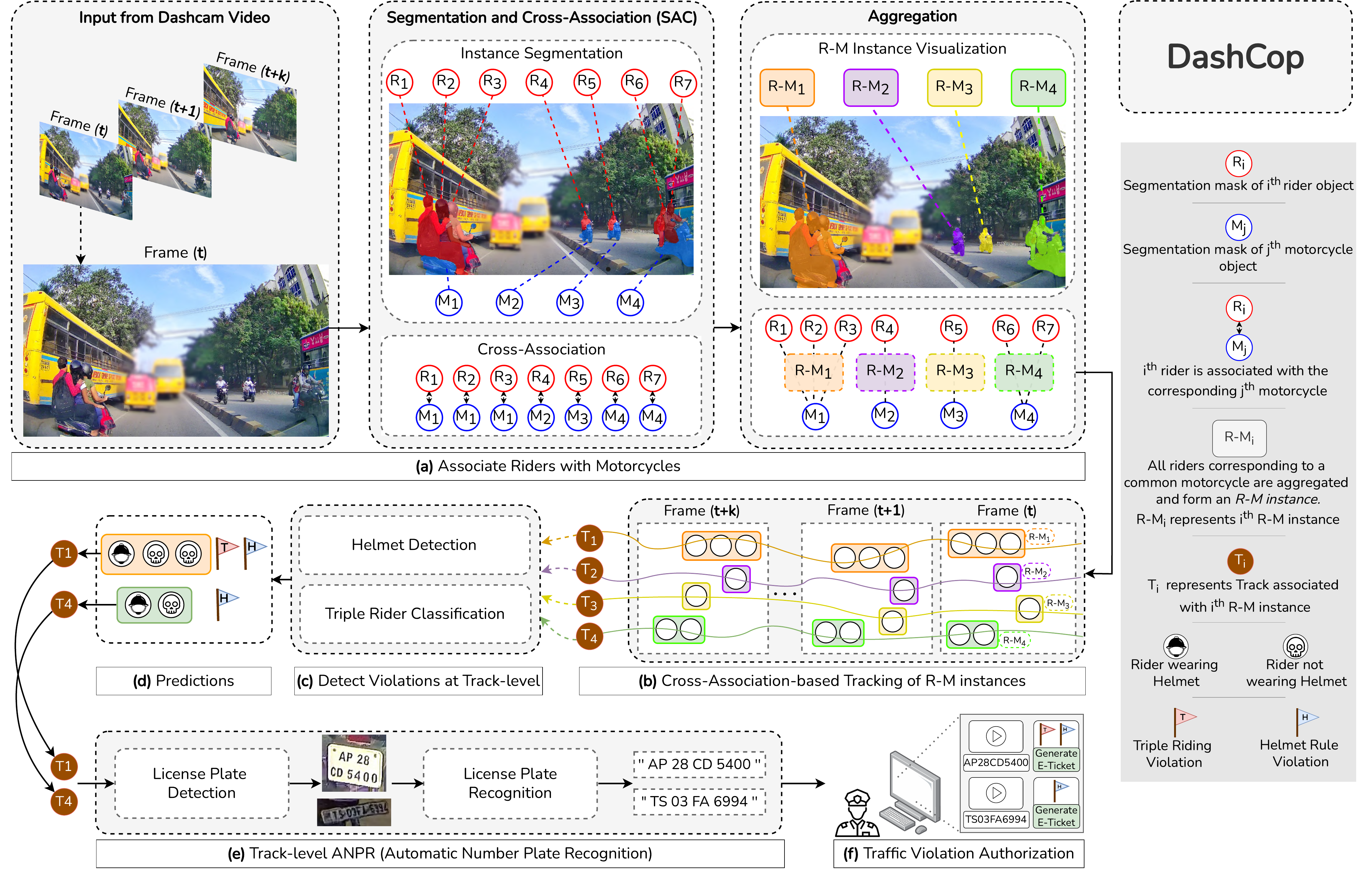}
   \caption{\textbf{Our Method:} (a) Dashcam video frames are input to our Segmentation and Cross-Association (SAC) module, which associates riders with their motorcycles to form rider-motorcycle (R-M) instances. (b) These R-M instances are then fed into a cross-association-based tracker, which robustly tracks them across frames and outputs track information \(T_i\) for each R-M instance \(R\)-\(M_i\). (c) This track information is used by the violation module to predict triple riding and helmet rule violations. (d) In the illustration, R-M instance \(R\)-\(M_1\) is flagged for both triple riding (red flag) and helmet rule violations, while \(R\)-\(M_4\) is flagged for a helmet rule violation (blue flag). (e) The track information \(T_1\) and \(T_4\), corresponding to \(R\)-\(M_1\) and \(R\)-\(M_4\) respectively, are fed into the ANPR module, which detects and reads the license plates. (f) Finally, the generated E-ticket, with supporting evidence, is shared with traffic authorities for authorization.}
\label{fig:method}
\end{center}
\end{figure*}


Figure~\ref{fig:method} provides an overview of our proposed end-to-end E-ticket generation method for detecting triple riding and helmet rule violations. As the first step, each frame from the dashcam video is processed by our novel Segmentation and Cross-Association (SAC) module (\cref{sec:SAC}). This module generates segmentation masks of riders and motorcycles using a YOLO-v8 framework~\cite{Jocher_Ultralytics_YOLO_2023}. We introduce novel extensions to the standard YOLO-v8 framework, which predicts the association of rider and motorcycle masks. The aggregation of all the riders on the same motorcycle comprises an R-M instance (see $R$-$M_1$, $R$-$M_2$, $\ldots$ in \cref{fig:method} (a)). 

Next, our novel cross-association-based tracker module (\cref{sec:rmtracker}) operates across video frames and assigns unique track IDs to R-M instances (\cref{fig:method} (b)). Each tracked R-M instance is then processed to detect potential traffic violations. For this, the region-of-interest (ROI) image crop of each R-M instance is fed into the violation module (\cref{fig:method} (c)), which includes a helmet detection module (\cref{sec:helmetdet}) and a triple rider classification module (\cref{sec:triplerideclassification}). The per-frame predictions of these modules are consolidated to indicate the presence of a potential triple riding violation or a helmet rule violation for each track. 

The tracks with detected traffic violations are further processed by the ANPR module. This module first detects the license plates within the R-M instance and then recognizes the license plate characters (\cref{sec:licensedetocr}, \cref{fig:method} (e)). Finally, the generated E-ticket (license plate data, along with the corresponding violation information and the R-M track frames) is made available to traffic authorities for verification and E-ticket generation (\cref{fig:method} (f)).


\subsection{Segmentation and Cross-Association (SAC)}
\label{sec:SAC}

\begin{figure*}
\begin{center}
\includegraphics[width=0.85\linewidth]{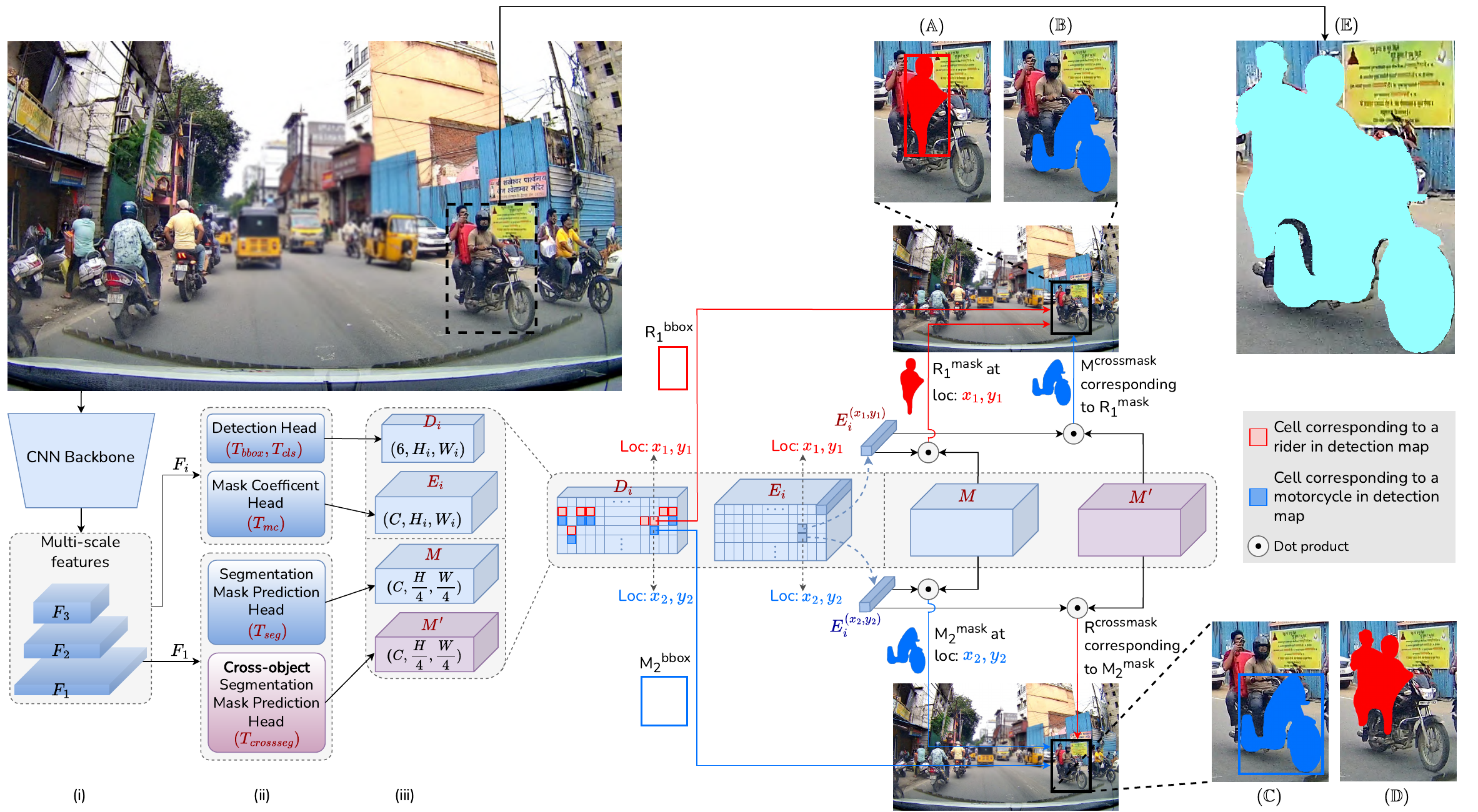}
   \caption{\textbf{Segmentation and Association (SAC):} The input frame (top-left) is processed by the SAC module to predict rider locations ($\mathbb{A}$), motorcycle locations ($\mathbb{C}$), and cross-object segmentation masks ($\mathbb{B,D}$). ($\mathbb{E}$) visualizes the R-M instance. In the architecture diagram, the blue blocks are from YOLO-v8~\cite{Jocher_Ultralytics_YOLO_2023}. The purple blocks are our novel addition which enable association. Refer to Sec.~\ref{sec:SAC} for details.}
\label{fig:sac}
\end{center}
\end{figure*}

Given a frame $I \in \mathbb{R}^{3 \times H \times W}$ from a video sequence, we first use a CNN backbone~\cite{Jocher_Ultralytics_YOLO_2023} to obtain multi-scale features $F_i, i =1, 2, 3, \ldots$ (\cref{fig:sac} (i)), where $i$ denotes the scale index. The extracted feature maps are fed to the detection head and mask coefficient head (\cref{fig:sac} (ii)). For each $F_i$, the detection head outputs $D_i = [B_i, K_i]$, where $B_i$ is the bounding box coordinate prediction map of shape $(4, H_i, W_i)$, and $K_i$ is the class scores prediction map of shape $(2, H_i, W_i)$, representing the object classes `motorcycle' and `rider'. 




\textbf{Instance segmentation:} The mask coefficient head outputs an intermediate mask coefficient embedding map $E_i$ of shape $(C, H_i, W_i)$, where each $E_i^{xy}$ is a vector along the channel that encodes an instance's representation at cell $(x,y)$ on $i^{th}$ scale. A parallel branch processes the high-resolution feature map $F_1$ to output 
$M$ of shape $(C, \frac{H}{4}, \frac{W}{4})$. To predict the instance segmentation mask for each spatial location $(x,y)$ in the $F_i$, we take the dot product between $M$ and $E^{xy}_i$ 
\begin{equation}
  S_{i}^{xy} = M\cdot E^{xy}_i
  \label{eq:sixth}
\end{equation}
where $S_{i}^{xy}$ is of shape $(\frac{H}{4}, \frac{W}{4})$.

Intuitively, $E^{xy}_i$ represents a query embedding, whereas $M$ represents the key embeddings. $E^{xy}_i$ attends strongly to spatial locations where the corresponding object is present in $M$. Consequently, the key locations which have a high similarity with $E^{xy}_i$ become a part of the final segmentation mask $S_i^{xy}$. 

\textbf{Cross-Association:} 
Traditional instance segmentation treats riders and motorcycles as independent entities. To model the association between riders and their corresponding motorcycles, we introduce a novel cross-object segmentation mask prediction head. This head follows the same segmentation mechanism as that of instance segmentation described above, but with a key modification. The query $E^{xy}_i$ for a rider instance is now encouraged to attend to location(s) where the rider's motorcycle is present (see $M^{crossmask}$ in \cref{fig:sac}). Conversely, the query for the motorcycle is encouraged to attend to \textit{all} locations corresponding to all the riders on the motorcycle ($R^{crossmask}$ in \cref{fig:sac}). 
The cross-object segmentation mask is predicted as
\begin{equation}
  A_{i}^{xy} = M' \cdot E^{xy}_i
  \label{eq:eighth}
\end{equation}
where $A_{i}^{xy}$ is of shape $(\frac{H}{4}, \frac{W}{4})$.




During inference, Non-Maximum Suppression (NMS) is applied to identify spatial locations in $F_i$ where the class scores exceed a predefined threshold. Segmentation and cross-object segmentation masks are then retrieved for these spatial locations using \cref{eq:sixth,eq:eighth}, resulting in a set of masks and cross-object masks for both riders and motorcycles. The resulting set of bounding boxes, $B^k$, segmentation masks, $S^k$, and cross-object segmentation masks, $A^k$, are expressed as $[(B_r^1, S_r^1, A_r^1), \ldots, (B_r^N, S_r^N, A_r^N)]$ for riders and $[(B_m^1, S_m^1, A_m^1), \ldots, (B_m^M, S_m^M, A_m^M)]$ for motorcycles. The association score $a_{k, l}$ between the $k$-th rider and $l$-th motorcycle detections is computed as

\begin{equation}
\makebox[0.9\columnwidth]{%
    \resizebox{0.9\columnwidth}{!}{$
    a_{k, l} = \frac{1}{2}( \text{IoU}(A_r^k, S_m^l) + \text{IoU}(S_r^k, A_m^l \cdot \text{mask}(B_r^k)) )
    $}
}
\label{eq:assoc-score}
\end{equation}

where $\text{mask}(B)$ is a function that takes in a bounding box and returns a binary mask with values inside the box set to 1 and outside set to 0.



\subsection{Rider-Motorcycle Association-based Tracking}
\label{sec:rmtracker}

\textit{SORT} is a popular, efficient, and robust family of trackers~\cite{bewley2016simple, wojke2017simple, botsort, bytetrack, ocsort, hybrid}. Tracking is performed via a bipartite matching between the existing tracks and current detections. This is done using a score matrix that represents the likelihood of assigning the $i^{th}$ track to the $j^{th}$ detection. This is modeled by assigning a binary variable $a_{ij} \in \{0, 1\}$ to each combination of the current tracks and the detections. If $s_{ij}$ denotes the score of assigning the $i^{th}$ track to the $j^{th}$ detection, the goal is to maximize
$a^{*} = \underset{a_{ij} \in \{0, 1\}}{\arg\max}\sum_{i=1}^{i=N}\sum_{j=1}^{j=M} a_{ij}\cdot{s_{ij}}
$ subject to the constraints

\begin{align}
 \sum_{i=1}^{i=N} a_{ij} \leqslant 1 \; \forall \; j \in \{1 \dots M\} 
 \label{eq:constraint0} \\
 \sum_{j=1}^{j=M} a_{ij} \leqslant 1 \; \forall \; i \in \{1 \dots N\}
\label{eq:constraint1}
\end{align}
 
These constraints ensure that each detection can be matched to only one track, and vice versa. 
The score $s_{ij}$ is a function of a motion model (\eg a Kalman filter) and an appearance model (\eg object re-identification).

\textbf{Cross-AssociationSORT:} In contrast to tracking object classes independently, our method aims to track both `rider' and `motorcycle' instances \textit{jointly}, while keeping track of the evolving associations between members of these object classes. This approach offers two key advantages: firstly, it improves tracking accuracy for each object, and secondly, it establishes robust associations among R-M instances over time.




To implement cross-association-based tracking, we utilize R-M association scores defined earlier (see \cref{eq:assoc-score}). Let the rider tracks up to time $t$ be represented by $T^r_t = \{r_1, r_2, \dots r_{N_r^t}\}$ and the motorcycle tracks by $T^m_t = \{m_1, m_2, \dots m_{N_m^t}\}$, where $N_r^t$ and $N_m^t$ denote the number of rider and motorcycle tracks at time $t$, respectively. The observations/detections at time $t+1$ are denoted by the set $D^r_{t+1}$ and $D^m_{t+1}$. Let the set of all possible assignments for the rider and motorcycle class be represented by

\begin{align}
C^r = \{(r_i, d_j) \; | \; r_i \in T^r_t \; \& \; d_j \in D^r_{t+1}\} \\
C^m = \{(m_i, d_j) \; | \; m_i \in T^m_t \; \& \; d_j \in D^m_{t+1}\}    
\end{align}

We define a binary variable $t_i^r$, for the $i^{th}$ element in $C^r$, represented by $C^r_i$ denoting its possibility of being a rider track at time $t+1$. Similarly $t_j^m$ is another binary variable defined for $C^m_j$.
Additionally, we define a likelihood score based on R-M association alone for every pair of combinations, one from $C^r$ and the other from $C^m$. For the $i^{th}$ rider hypothesis in $C^r$ and the $j^{th}$ motorcycle hypothesis in $C^m$, we define a binary variable $e_{ij}$ with an associated score of $a_{ij}$, which reflects the likelihood of association of these two hypotheses.
If and only if both $C^r_i$ and $C^m_j$ are chosen, only then there is a possibility of $e_{ij} = 1$, implying the association of the two tracks. This can be represented as

\begin{align}
e_{ij}(1 - t^r_i \cdot t^m_j) = 0 \; \forall \; i,j    
\label{eq:constrain2}
\end{align}



Additionally, to ensure that a rider can only be associated with one motorcycle, we impose the constraint

\begin{align}
\sum_{j=1}^{j=|C^m|} e_{ij} \leqslant 1 \; \forall \; i
\label{eq:constrain3}
\end{align}

In our modified formulation, we maximize 

\footnotesize  

\begin{equation}
\begin{aligned}
(t^{r*}, t^{m*}, e^{*}) =  \underset{\{t^r_i, t^m_j, e_{ij}\} \in \{0, 1\}}{\arg\max} \; &\left( \lambda_1 \sum_{i=1}^{|C_r|} t_{i}^{r} s_i^r + \lambda_2 \sum_{j=1}^{|C_m|} t_{j}^{m} s_j^m \right. \\
&\left. + \lambda_3 \sum_{i=1}^{|C_r|} \sum_{j=1}^{|C_m|} e_{ij} a_{ij} \right)
\end{aligned}
\end{equation}

\normalsize  

subject to the aforementioned (\cref{eq:constrain2}, \cref{eq:constrain3}) and the original (\cref{eq:constraint0}, \cref{eq:constraint1}) constraints. Here, $s_i^r$ is the score for $i^{th}$ rider hypothesis and $s_j^m$ is the score for $j^{th}$ motorcycle hypothesis. The values obtained by the variables, $t_i^r$ and $t_j^m$ after the optimization will reflect the next set of tracks at time $t+1$, and will be associated based on the value of $e_{ij}$.


\subsection{Traffic Violation Detection}

\subsubsection{Helmet Detection Module}
\label{sec:helmetdet}

Each video frame is processed by a helmet detection module, trained to classify `helmet' and `no-helmet'. The module outputs detections for each frame. These detection boxes are then matched with corresponding rider tracks. For each rider track, we determine the predominant class (`helmet' or `no-helmet') across multiple frames using majority voting.

\subsubsection{Triple Rider Classification Module}
\label{sec:triplerideclassification}

The input to the triple riding classification model consists of R-M instance ROI crops output by the cross-association-based tracking module. The classifier is designed as a learnable convolutional layer on the frozen mask coefficient head of the SAC module (\cref{sec:SAC}), followed by a learnable linear layer. The model classifies the rider count as `single', `double', `triple' or `none'. During training, the backbone of SAC is frozen. An R-M track is flagged as a triple riding violation if at least one instance ROI in the track is classified as `triple riding'.

\subsection{License Plate Recognition and E-ticket Generation}
\label{sec:licensedetocr}


The license plate detection model is trained to identify license plate regions within the R-M instance crops. For tracks with detected violations, the ROIs from all frames are passed through the license plate detector. Detected license plate crops are then input to an OCR model. A lightweight OCR architecture~\cite{crnn} is trained on a dataset of real-world, augmented, and synthetic license plates. The license plate numbers detected across multiple frames for a given R-M instance track are consolidated using a majority vote, where the most frequently detected license plate number is selected and linked to the corresponding R-M instance track for E-ticket generation.

\section{Training and Implementation Details}

\subsection{Segmentation and Cross-Association (SAC)} 

The bounding box and class predictions are matched to the ground truth using the Task-Alignment Learning (TAL) approach~\cite{feng2021tood}. The matched boxes are evaluated against the ground truth boxes using the Complete IoU (CIOU) loss~\cite{ciou} and the Distribution Focal Loss (DFL)~\cite{dfl}. For classification, Binary Cross-Entropy (BCE) loss is used. For segmentation and cross-object segmentation, a per-pixel BCE loss is used.
The overall loss function is formulated as $\mathcal{L}_{SAC} = \lambda_1 L_{CIOU} + \lambda_2  L_{DFL} + \lambda_3 L_{cls} + \lambda_4 L_{seg} + \lambda_5 L_{crossseg}$ where $\lambda_1=7.5, \lambda_2=1.5, \lambda_3=0.5, \lambda_4=7, \lambda_5=7$ are hyperparameters for balancing multi-task learning objectives. The model is trained for 200 epochs with a learning rate of 0.01, using the Adam optimizer on 4 NVIDIA GeForce RTX 3080 Ti GPUs over 4 days.

\subsection{Rider-Motorcycle Association-based Tracking}

We use the pre-trained Re-Identification (ReID) model SBS-R50~\cite{he2020fastreid} for appearance modeling and a Kalman filter for motion modeling. The optimization process is handled using the Gurobi solver~\cite{gurobi}.
Segmentation masks from SAC module are stored for a fixed sized buffer of $k=3$ time steps. Using these, for computing $a_{ij}$, we calculate the sum of association scores for the rider and motorcycle detection for the $k$ time steps, i.e. $\sum_{t=T-k}^{T} \text{A}(C^r_{i^t}, C^m_{j^t}) - \theta$, where $C^r_{i^t}$ and $C^m_{j^t}$ represent the rider and motorcycle detections at the $t^{th}$ time step, and $\theta$ is a threshold set to $0.5$. The function $A$ is implemented using \cref{eq:assoc-score}. If there is no detection for either $C^r_{i^t}$ or $C^m_{j^t}$, then function $A$ returns a value of 0. If the association IDs of the tracks is already established and the association ID of the $i^{th}$ rider hypothesis is not equal to the association ID of the $j^{th}$ motorcycle hypothesis, then $a_{ij}$ is set to -$\infty$. We use the validation set of RideSafe-400 dataset for tuning the tracker's hyperparameters.

\subsection{Traffic Violation Detection}

The Helmet Detection module is based on the YOLOv8-x~\cite{Jocher_Ultralytics_YOLO_2023}, trained to classify two categories: `helmet' and `no-helmet'. The model is trained using full-resolution frames, with a dataset consisting of 37K helmet and 23K no-helmet bounding boxes. The training process spans 70 epochs, using an Adam optimizer with a learning rate of 0.01, on four NVIDIA GeForce RTX 3080 Ti GPUs for one day.

\subsection{License Plate Detection and Recognition}

A YOLOv8-x model~\cite{Jocher_Ultralytics_YOLO_2023} is trained for license plate detection. To improve performance during both training and inference, multi-line license plates of motorcycles are converted into a single-line format. License plate recognition is performed using an OCR model~\cite{crnn}, which is trained on real-world, augmented, and synthetic license plate data. 
Further details can be found in Supplementary.

\section{Experiments and Results}

We conducted experiments using a dataset of 400 videos, divided into 200 videos for training, 100 videos for validation, and 100 videos for final evaluation (test).

\subsection{System-Level Performance (E-ticket Generation)}
\label{sec:sys-level}

\begin{table}[htbp]
    \centering
    \resizebox{\columnwidth}{!}{
        \begin{tabular}{|T|T|T|D|D|D|R|R|E|E|E|}
            \hline
            \multicolumn{3}{|T|}{Traffic Violation} & \multicolumn{3}{D|}{LP Detection} & \multicolumn{2}{R|}{LP Recognition} & \multicolumn{3}{E|}{\textbf{E-ticket}}\\
            \hline
            TP & FP & FN & TP & FP & FN & Correct & Incorrect & \textbf{TP} & \textbf{FP} & \textbf{FN} \\
            \hline
            \hline
            \checkmark & & & \checkmark & & & \checkmark & & \checkmark & & \\
            \hline
            \checkmark & & & \checkmark & & & & \checkmark & & \checkmark & \\
            \hline
            & \checkmark & & \checkmark & & & & \checkmark & & \checkmark & \\
            \hline
            & \checkmark & & \checkmark & & & \checkmark & & & \checkmark & \\
            \hline
            \checkmark & & & & \checkmark & & & \checkmark & & \checkmark & \\
            \hline
            & \checkmark & & & \checkmark & & & \checkmark & & \checkmark & \\
            \hline
            & & \checkmark & & & & & & & & \checkmark \\
            \hline
            \checkmark & & & & & \checkmark & & & & & \checkmark \\
            \hline
        \end{tabular}
    }
    \caption{\textbf{Evaluation criteria for E-ticket generation:} For a given R-M track, the labelling (TP, FP, FN) at various stages of system pipeline is used to determine the final E-ticket level label of the track. For e.g., a True Positive (TP) prediction at all stages of the pipeline is considered a TP for E-ticket generation system. \textit{Correct} means that all the characters of the license plate number match the ground-truth. Refer to \cref{sec:sys-level} for details.}
    \label{tab:sys-level-metrics}
\end{table}

\begin{table}[!t]
    \centering
    \resizebox{\columnwidth}{!}{
        \begin{tabular}{l|ccc}
            \hline
            \textbf{E-ticket System Output Level} & \textbf{Precision \%} & \textbf{Recall \%} & \textbf{F1-score \%} \\ \hline
             Automatic & $\;$84.21 & 63.16 &  72.18 \\
             Human-In-The-Loop & 100.00 & 69.57 &  82.05 \\
            \hline
        \end{tabular}
    }
    \caption{Overall E-ticket system performance (\cref{sec:sys-level}).}
    \label{tab:sys-level}
\end{table}

\begin{figure}[!t]
  \centering  
   \includegraphics[width=\linewidth]{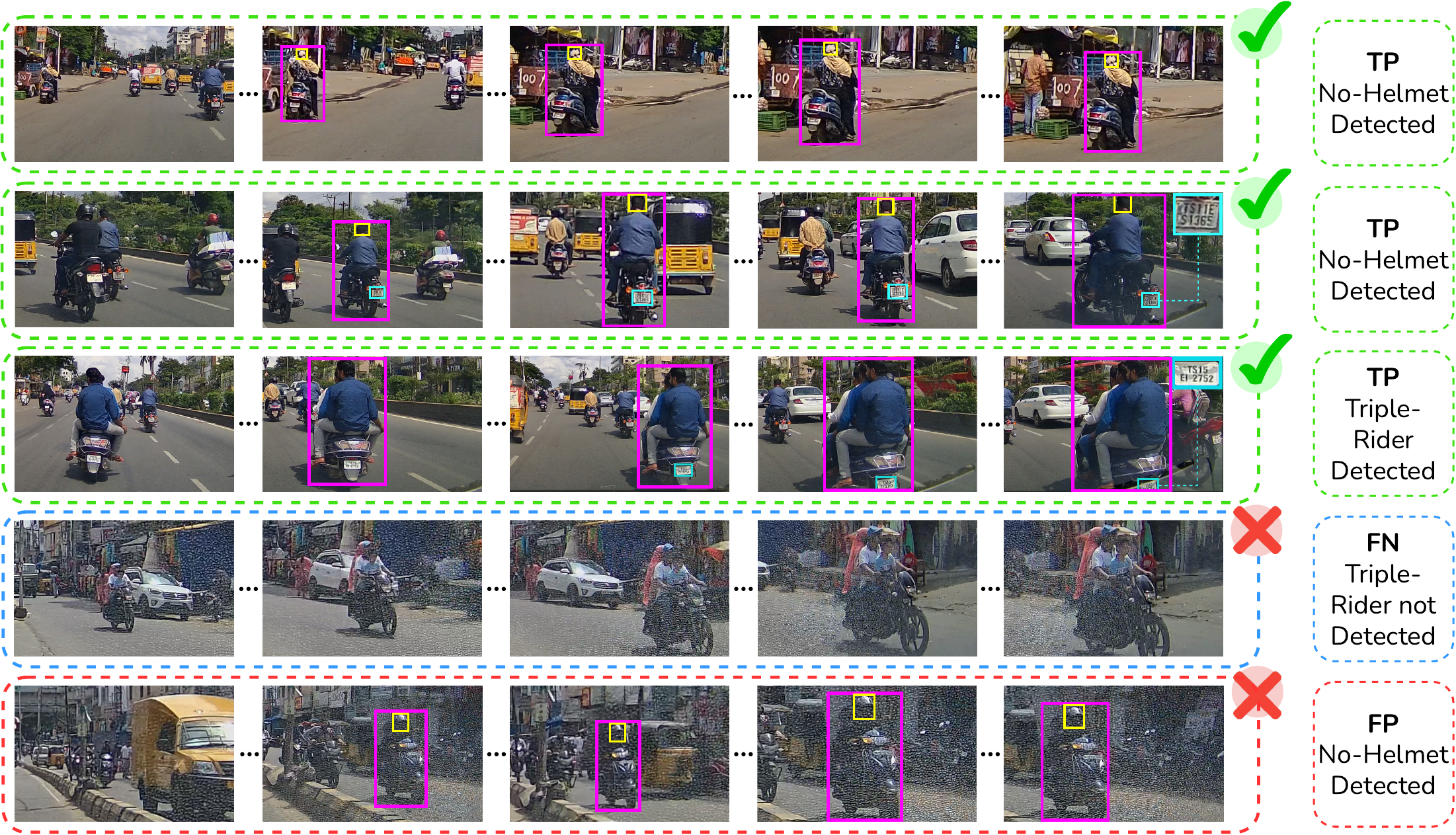}
   \caption{Qualitative results of our system and failure cases. R-M instances are indicated by pink bounding boxes. Yellow indicates a No-Helmet detection.}   
   \label{fig:qualitative-sys}
\end{figure}

We evaluate the system using correctly tracked R-M instances. An R-M track is considered a true positive for E-Ticket generation if the system accurately predicts the traffic violation (\eg no-helmet or triple riding) and correctly detects and recognizes the license plate. False positives and false negatives are calculated similarly (see \cref{tab:sys-level-metrics}). As shown in \cref{tab:sys-level}, the automated system achieves an F1-score of 72.18\%, reflecting good overall performance. In practice, traffic enforcement personnel review E-Tickets and corresponding evidence before issuance. This human-in-the-loop approach eliminates false positives, raising the F1-score to 82.05\%, improving system reliability.
Some examples of system performance can be seen in \cref{fig:qualitative-sys}. Empirical observations indicate that the system struggles with detecting triple riders in scenarios involving significant glare or motion blur, and helmet detection suffers from occlusions due to objects or vehicles, leading to false negatives. Furthermore, the triple riding classifier misclassified distant occluded R-M instances, contributing to false positives. License plate recognition is most accurate when R-M instances move in the same direction as the ego vehicle. However, the recognition rate drops for R-M instances approaching from the opposite lane.

\subsection{Violation Detection}
\label{sec:violation-results}

\noindent \textit{Helmet Violation:} \cref{tab:hnh_results} presents comparative evaluation results for helmet violation detection at the rider track level are. Our YOLOv8-x-based frame-level approach achieves the best F1-score, striking a good balance between precision and recall. The results show that frame-level context is marginally more effective compared to detecting helmet violations at R-M instance crop level. Compared to prior works, YOLOv8-x based methods show better balance across different ROI extraction methods.

\begin{table}[!t]
    \centering   
    \resizebox{\columnwidth}{!}{
        \begin{tabular}{ll|ccc}
            \hline
            \textbf{Method} & \textbf{ROI Extraction Approach} & \textbf{Precision \%} & \textbf{Recall \%} & \textbf{F1-score \%} \\
            \hline
            Goyal \etal ~\cite{goyal2022detecting} & R-M Instance & 51.42 & 32.34 & 39.71\\
            \hline
            Cui \etal \cite{hnh_comp} & Frame & \textbf{76.34} & 39.11 & 51.72\\
            \hline
            \textbf{\multirow{3}{*}{YOLOv8-x \cite{Jocher_Ultralytics_YOLO_2023}}} & Rider Instance & 67.15 & 53.57 &  59.60 \\
            \cline{2-5}
             & R-M Upper Instance & 67.98 & 58.11 &  62.66 \\
             \cline{2-5}
             & R-M Instance& 73.48 & 57.85 & 64.73 \\
            \cline{2-5}
             \textbf{Ours} & \textbf{Frame} & 68.75 & \textbf{62.78} &  \textbf{65.63} \\
            \hline
        \end{tabular}
    }
    \caption{Comparative evaluation of helmet violation detection.}
    \label{tab:hnh_results}
\end{table}

\noindent \textit{Triple Riding Violation:} The results in \cref{tab:tripleriding_results} indicate that our R-M instance-based classification approach yields the best performance. Methods based on counting rider or helmet detections are less robust to occlusions. Combining our classifier with rider-counting greatly reduces false negatives. However, it increases the false positive rate.

\begin{table}
    \centering    
    \resizebox{\columnwidth}{!}{
            \begin{tabular}{l|ccc}
            \hline
                \textbf{Method} & \textbf{Precision \%} & \textbf{Recall \%} & \textbf{F1-score \%} \\
            \hline
             Goyal \etal~\cite{goyal2022detecting} & 32.30 & 01.46 &  02.79 \\
             \hline
            Rider-Count & 54.86 & 14.60 &  23.07 \\
            Helmet Detector & 65.92 & 59.55 &  62.57 \\
            
             TR Classifier + Rider-Count & 59.51 & \textbf{75.28} &  66.40 \\
             TR Classifier + Helmet Detector & 57.88 & 43.82 &  49.88 \\
             \textbf{TR Classifier} & \textbf{69.22} & 66.29 &  \textbf{67.72} \\
             \hline
            \end{tabular}
         }
\caption{Comparative evaluation of triple riding violation.}
    \label{tab:tripleriding_results}    
\end{table}

\subsection{Rider-Motorcycle Detection and Association}
\label{sec:rmassoc_results}

We define the association score as the percentage of correctly associated R-M instances. Our SAC module outperforms association methods involving geometric heuristics based on bounding box overlapping (\cref{tab:rmassoc_results}). The trapezium regressor~\cite{goyal2022detecting} does not take into account the visual information, and directly operates on the coordinates of the bounding boxes present. In contrast, our approach operates directly in the image space and learns a more robust association between the classes.

\begin{table}
    \centering
    \resizebox{\columnwidth}{!}{
        \begin{tabular}{ll|c}
        \hline
            \textbf{Detection Method} & \textbf{R-M Association} & \textbf{Association Score \%} \\
         \hline
            \multirow{2}{*}{Goyal \etal~\cite{goyal2022detecting}} & IOU & 58.50  \\
             &  Trapezium Regressor & 59.32  \\
             \hline
          \textbf{\multirow{3}{*}{Ours}} & IOU & 80.93  \\
           & Trapezium Regressor~\cite{goyal2022detecting} & 81.71  \\
           & \textbf{SAC Module} & \textbf{84.04}  \\
            \hline
        \end{tabular}
    }
\caption{Evaluation of rider-motorcycle association.}
    \label{tab:rmassoc_results}
\end{table}

\subsection{Rider-Motorcycle Instance Tracking}

To evaluate the tracking of R-M instances, we use the widely accepted HOTA\cite{luiten2021hota}, MOTA \cite{clear}, and IDF1\cite{idf1} metrics. We compare with various SORT trackers. From \cref{tab:trk_results}, we see that joint tracking of R-M instances results in better tracking performance across all metrics than independently tracking component objects (riders, motorcycles) and post-hoc aggregation.

\begin{table}[!t]
    \centering    
    \resizebox{\columnwidth}{!}{
    \begin{tabular}{l|ccccc}
        \hline
        \textbf{Tracker} & \textbf{HOTA} & & \textbf{MOTA} & & \textbf{IDF1} \\
        \hline
         ByteTrack\cite{bytetrack} & 56.86 & & 48.60 & & 60.04 \\
         DeepOCSORT\cite{ocsort} & 58.21 & & 50.78 & & 62.57 \\
         BotSORT\cite{botsort} & 58.12 & & 50.72 & & 63.09 \\
         HybridSORT\cite{hybrid} & 58.10 & & 50.66 & & 62.29 \\
         \textbf{Cross-AssociationSORT (Ours)} & \textbf{60.41} & & \textbf{51.96} & &  \textbf{64.58}\\
        \hline              
    \end{tabular}
    }
    \caption{Evaluation of tracking rider-motorcycle instances.}
    \label{tab:trk_results}
\end{table}

\subsection{License Plate OCR}
\label{sec:anpr_results}

For license plates, the metrics are Character Error Rate (CER) and accuracy (percentage of plates recognized correctly in their entirety).  From \cref{tab:lpr_results}, we see that CRNN~\cite{crnn} trained on a combination of synthetic and real data provides the best results. We found CNN-based approach to outperform more recent transformer-based approaches~\cite{li2023trocr} and commercial OCR ~\cite{googleVisionImage}.

\begin{table}
    \centering    
    \resizebox{\columnwidth}{!}{
        \begin{tabular}{cccc|cc}
            \hline
            \textbf{Method} & \textbf{Real} & \textbf{Synth} & \textbf{Aug} & \textbf{CER (↓)} & \textbf{Accuracy \%} \\
            \hline
            TrOCR \cite{li2023trocr} & \checkmark & \checkmark & \checkmark & 0.1691 & 16.41 \\
            \hline
             GoogleOCR \cite{googleVisionImage} &  &  &  & 0.2698 & 33.34 \\
            \hline
            \multirow{5}{*}{CRNN \cite{crnn}} 
            & \checkmark & & & 0.0507 & 71.42 \\
            & \checkmark & & \checkmark & 0.0602 & 71.43 \\
            & \checkmark & \checkmark & \checkmark & 0.0503 & 80.95 \\
            & \checkmark & \checkmark & & \textbf{0.0400} &  \textbf{85.71}\\
            \hline
        \end{tabular}
    }
    \caption{Evaluation of license plate recognition methods.}
    \label{tab:lpr_results}
\end{table}

\section{Conclusion}

DashCop is a comprehensive end-to-end dashcam-video based system for automated E-ticket generation targeting traffic violations in motorized two-wheelers. The key innovations include the SAC module for precise rider-motorcycle (R-M) association, Cross-AssociationSORT for robust R-M instance tracking, and system-level performance evaluation criteria. The large-scale RideSafe-400 video dataset, featuring multi-level and multi-label annotations, is another key contribution. Validated on this dataset, DashCop shows reliable E-ticket generation capabilities, presenting a promising solution for enhancing traffic safety through automated enforcement and deterrence.

\noindent \textit{Societal concerns:} While enhancing road safety, the proposed technology raises privacy concerns. To mitigate these risks, we adhere to relevant legal and ethical standards governing surveillance-based public traffic management systems. Our system is designed to relay only relevant information about detected traffic violations to authorities. Thus, we minimize the risks associated with sharing  complete footage indiscriminately. Also, we employ anonymization techniques to obscure identifiable information (\eg faces). Even for license plates, only the instances corresponding to traffic violations are shown. The data is stored securely, with access restricted to authorized personnel. 

\noindent\textbf{Acknowledgements:} This work is supported by IHub-Data, IIIT Hyderabad.


\begin{thebibliography}{10}\itemsep=-1pt

\bibitem{aboah2023real}
Armstrong Aboah, Bin Wang, Ulas Bagci, and Yaw Adu-Gyamfi.
\newblock Real-time multi-class helmet violation detection using few-shot data sampling technique and yolov8.
\newblock In {\em Proceedings of the IEEE/CVF conference on computer vision and pattern recognition}, pages 5349--5357, 2023.

\bibitem{markntel2023}
MarkNtel Advisors.
\newblock Global two wheeler market research report: Forecast (2023-2028), 2023.

\bibitem{botsort}
Nir Aharon, Roy Orfaig, and Ben-Zion Bobrovsky.
\newblock Bot-sort: Robust associations multi-pedestrian tracking.
\newblock {\em arXiv preprint arXiv:2206.14651}, 2022.

\bibitem{ariffin2021exploring}
AH Ariffin, MS Solah, A Hamzah, MS Ahmad, MA~Mohamad Radzi, ZH Zulkipli, AS Salleh, and Z~Mohd Jawi.
\newblock Exploring characteristics and contributory factors of road crashes and near misses recorded via dashboard camera.
\newblock {\em Journal of the Society of Automotive Engineers Malaysia}, 5(2):292--305, 2021.

\bibitem{barreto2021object}
Fabian Barreto, Nesline~D Almeida, Premraj Nadar, Ravneet Kaur, and Sanjana Khairnar.
\newblock Object detection and e-challan generation systems for traffic violation: A review.
\newblock {\em International Journal of Research in Engineering, Science and Management}, 4(5):54--57, 2021.

\bibitem{clear}
Keni Bernardin and Rainer Stiefelhagen.
\newblock Evaluating multiple object tracking performance: the clear mot metrics.
\newblock {\em EURASIP Journal on Image and Video Processing}, 2008:1--10, 2008.

\bibitem{bewley2016simple}
Alex Bewley, Zongyuan Ge, Lionel Ott, Fabio Ramos, and Ben Upcroft.
\newblock Simple online and realtime tracking.
\newblock In {\em 2016 IEEE international conference on image processing (ICIP)}, pages 3464--3468. IEEE, 2016.

\bibitem{bouhayane2023swin}
Ayyoub Bouhayane, Zakaria Charouh, Mounir Ghogho, and Zouhair Guennoun.
\newblock A swin transformer-based approach for motorcycle helmet detection.
\newblock {\em IEEE Access}, 11:74410--74419, 2023.

\bibitem{carion2020end}
Nicolas Carion, Francisco Massa, Gabriel Synnaeve, Nicolas Usunier, Alexander Kirillov, and Sergey Zagoruyko.
\newblock End-to-end object detection with transformers.
\newblock In {\em European conference on computer vision}, pages 213--229. Springer, 2020.

\bibitem{chandravanshi2021automated}
Shubham~Kumar Chandravanshi, Hirva Bhagat, Manan Darji, and Himani Trivedi.
\newblock Automated generation of challan on violation of traffic rules using machine learning.
\newblock {\em International Journal of Science and Research (IJSR)}, 10(3):1157--1162, 2021.

\bibitem{charran2022two}
R~Shree Charran and Rahul~Kumar Dubey.
\newblock Two-wheeler vehicle traffic violations detection and automated ticketing for indian road scenario.
\newblock {\em IEEE Transactions on Intelligent Transportation Systems}, 23(11):22002--22007, 2022.

\bibitem{chaturvedi2023detection}
Pooja Chaturvedi, Kruti Lavingia, and Gaurang Raval.
\newblock Detection of traffic rule violation in university campus using deep learning model.
\newblock {\em International Journal of System Assurance Engineering and Management}, 14(6):2527--2545, 2023.

\bibitem{chen2024effective}
Yunliang Chen, Wei Zhou, Zicen Zhou, Bing Ma, Chen Wang, Yingda Shang, An Guo, and Tianshu Chu.
\newblock An effective method for detecting violation of helmet rule for motorcyclists.
\newblock In {\em Proceedings of the IEEE/CVF Conference on Computer Vision and Pattern Recognition}, pages 7085--7090, 2024.

\bibitem{choudhari2023traffic}
Rutvik Choudhari, Shubham Goel, Yash Patel, and Sunil Ghane.
\newblock Traffic rule violation detection using detectron2 and yolov7.
\newblock In {\em 2023 World Conference on Communication \& Computing (WCONF)}, pages 1--7. IEEE, 2023.

\bibitem{cui2023effective}
Shun Cui, Tiantian Zhang, Hao Sun, Xuyang Zhou, Wenqing Yu, Aigong Zhen, Qihang Wu, and Zhongjiang He.
\newblock An effective motorcycle helmet object detection framework for intelligent traffic safety.
\newblock In {\em Proceedings of the IEEE/CVF Conference on Computer Vision and Pattern Recognition}, pages 5469--5475, 2023.

\bibitem{hnh_comp}
Shun Cui, Tiantian Zhang, Hao Sun, Xuyang Zhou, Wenqing Yu, Aigong Zhen, Qihang Wu, and Zhongjiang He.
\newblock An effective motorcycle helmet object detection framework for intelligent traffic safety.
\newblock In {\em Proceedings of the IEEE/CVF Conference on Computer Vision and Pattern Recognition}, pages 5470--5476, 2023.

\bibitem{duong2023helmet}
Viet~Hung Duong, Quang~Huy Tran, Huu Si~Phuc Nguyen, Duc~Quyen Nguyen, and Tien~Cuong Nguyen.
\newblock Helmet rule violation detection for motorcyclists using a custom tracking framework and advanced object detection techniques.
\newblock In {\em Proceedings of the IEEE/CVF Conference on Computer Vision and Pattern Recognition}, pages 5380--5389, 2023.

\bibitem{feng2021tood}
Chengjian Feng, Yujie Zhong, Yu Gao, Matthew~R Scott, and Weilin Huang.
\newblock Tood: Task-aligned one-stage object detection.
\newblock In {\em 2021 IEEE/CVF International Conference on Computer Vision (ICCV)}, pages 3490--3499. IEEE Computer Society, 2021.

\bibitem{yahoo2024}
Yahoo Finance.
\newblock Global dashboard camera market analysis 2023-2030, 2024.

\bibitem{ghosh2023deep}
Rajdeep Ghosh, Akash Kotal, Raja Karmakar, and Krittika Das.
\newblock A deep learning-based smart helmet detection and notification with e-challan issuance for traffic fines.
\newblock In {\em 2023 IEEE 4th Annual Flagship India Council International Subsections Conference (INDISCON)}, pages 1--6. IEEE, 2023.

\bibitem{googleVisionImage}
Vision ai: Image \& visual ai tools --- cloud.google.com.
\newblock \url{https://cloud.google.com/vision?hl=en}.

\bibitem{goyal2022detecting}
Aman Goyal, Dev Agarwal, Anbumani Subramanian, CV Jawahar, Ravi~Kiran Sarvadevabhatla, and Rohit Saluja.
\newblock Detecting, tracking and counting motorcycle rider traffic violations on unconstrained roads.
\newblock In {\em Proceedings of the IEEE/CVF conference on computer vision and pattern recognition}, pages 4303--4312, 2022.

\bibitem{gurobi}
{Gurobi Optimization, LLC}.
\newblock {Gurobi Optimizer Reference Manual}, 2024.

\bibitem{he2020fastreid}
Lingxiao He, Xingyu Liao, Wu Liu, Xinchen Liu, Peng Cheng, and Tao Mei.
\newblock Fastreid: A pytorch toolbox for general instance re-identification.
\newblock {\em arXiv preprint arXiv:2006.02631}, 2020.

\bibitem{fortune2023}
Fortune~Business Insights.
\newblock Two-wheeler market size, share \& covid-19 impact analysis, by type (scooter, motorcycle, and moped), by technology (ice and electric), and regional forecast, 2023-2030, 2023.

\bibitem{fmi2022}
Future~Market Insights.
\newblock Dashboard camera market by technology, video quality, application \& region | forecast 2022 to 2032, 2022.

\bibitem{gmi2023}
Global~Market Insights.
\newblock Electric two-wheeler market size - by vehicle type (electric motorcycle, electric scooter, e-bikes, electric kick scooter), battery (sla, li-ion), motor placement (hub-motor, frame-mounted motor), motor power, motor speed \& global forecast, 2023 - 2032, 2023.

\bibitem{mordor2023}
Mordor Intelligence.
\newblock Global two-wheeler market size \& share analysis - growth trends \& forecasts up to 2029, 2023.
\newblock Last accessed: 2023.

\bibitem{Jocher_Ultralytics_YOLO_2023}
Glenn Jocher, Ayush Chaurasia, and Jing Qiu.
\newblock {Ultralytics YOLO}, Jan. 2023.

\bibitem{joy2023traffic}
Salna Joy, Ujwal~A Reddy, Rishab~C Lal, R Vinay, et~al.
\newblock Traffic rule violation recognition for two wheeler using yolo algorithm.
\newblock In {\em 2023 Second International Conference on Electronics and Renewable Systems (ICEARS)}, pages 1477--1480. IEEE, 2023.

\bibitem{kar2017real}
Gorkem Kar, Shubham Jain, Marco Gruteser, Fan Bai, and Ramesh Govindan.
\newblock Real-time traffic estimation at vehicular edge nodes.
\newblock In {\em Proceedings of the Second ACM/IEEE Symposium on Edge Computing}, pages 1--13, 2017.

\bibitem{kumar2023triple}
Nikhil Kumar, Gaurav~Kumar Sahu, M Ravi, Sahil Kumar, V Sukruth, and AN~Mukunda Rao.
\newblock Triple riding and no-helmet detection.
\newblock In {\em 2023 4th IEEE Global Conference for Advancement in Technology (GCAT)}, pages 1--6. IEEE, 2023.

\bibitem{kumari2024effcient}
Mrs J~RATNA KUMARI, NADENDLA BHAVANI, SHAIK THALIB, VATLURU CHARAN NAGA~SAI SURYA, and BATHULA Srikanth.
\newblock An effcient system for detecting traffic violations such as over speed, disregarding signals, and instances of triple riding.
\newblock {\em Turkish Journal of Computer and Mathematics Education (TURCOMAT)}, 15(1):104--108, 2024.

\bibitem{li2023trocr}
Minghao Li, Tengchao Lv, Jingye Chen, Lei Cui, Yijuan Lu, Dinei Florencio, Cha Zhang, Zhoujun Li, and Furu Wei.
\newblock Trocr: Transformer-based optical character recognition with pre-trained models.
\newblock In {\em Proceedings of the AAAI Conference on Artificial Intelligence}, volume~37, pages 13094--13102, 2023.

\bibitem{dfl}
Xiang Li, Chengqi Lv, Wenhai Wang, Gang Li, Lingfeng Yang, and Jian Yang.
\newblock Generalized focal loss: Towards efficient representation learning for dense object detection.
\newblock {\em IEEE Transactions on Pattern Analysis and Machine Intelligence}, 45(3):3139--3153, 2023.

\bibitem{luiten2021hota}
Jonathon Luiten, Aljosa Osep, Patrick Dendorfer, Philip Torr, Andreas Geiger, Laura Leal-Taix{\'e}, and Bastian Leibe.
\newblock Hota: A higher order metric for evaluating multi-object tracking.
\newblock {\em International Journal of Computer Vision}, 129:548--578, 2021.

\bibitem{daniel2024}
Dr. Shikha~Tiwari M.~Shaun~Daniel.
\newblock Advanced traffic monitoring and enforcement using yolov8.
\newblock {\em International Journal of Research Publication and Reviews}, 5(1):2070--2073, 2024.

\bibitem{ocsort}
Gerard Maggiolino, Adnan Ahmad, Jinkun Cao, and Kris Kitani.
\newblock Deep oc-sort: Multi-pedestrian tracking by adaptive re-identification.
\newblock In {\em 2023 IEEE International Conference on Image Processing (ICIP)}, pages 3025--3029. IEEE, 2023.

\bibitem{mallela2021detection}
Nikhil~Chakravarthy Mallela, Rohit Volety, and Nadesh RK.
\newblock Detection of the triple riding and speed violation on two-wheelers using deep learning algorithms.
\newblock {\em Multimedia Tools and Applications}, 80(6):8175--8187, 2021.

\bibitem{miah2024analyzing}
Md~Mamun Miah, Biton Chakma, and Kabir Hossain.
\newblock Analyzing the prevalence of and factors associated with road traffic crashes (rtcs) among motorcyclists in bangladesh.
\newblock {\em The Scientific World Journal}, 2024(1):7090576, 2024.

\bibitem{nandhakumar2023yolo}
RG Nandhakumar, K~Nirmala Devi, N Krishnamoorthy, S Shanthi, VR Pranesh, and S Nikhalyaa.
\newblock Yolo based license plate detection of triple riders and violators.
\newblock In {\em 2023 International Conference on Computer Communication and Informatics (ICCCI)}, pages 1--6. IEEE, 2023.

\bibitem{toi2019}
The~Times of India.
\newblock Just 72,000 traffic cops to manage 20 crore vehicles, 2019.

\bibitem{peng2022traffic}
Yan-Tsung Peng, Chen-Yu Liu, He-Hao Liao, Wei-Cheng Lien, and Gee-Sern~Jison Hsu.
\newblock Traffic violation detection via depth and gradient angle change.
\newblock In {\em 2022 IEEE 7th International Conference on Intelligent Transportation Engineering (ICITE)}, pages 326--330. IEEE, 2022.

\bibitem{straits2022}
Straits Research.
\newblock Dashboard camera market, 2022.

\bibitem{idf1}
Ergys Ristani, Francesco Solera, Roger Zou, Rita Cucchiara, and Carlo Tomasi.
\newblock Performance measures and a data set for multi-target, multi-camera tracking.
\newblock In {\em European conference on computer vision}, pages 17--35. Springer, 2016.

\bibitem{rocky2024review}
Arash Rocky, Qingming~Jonathan Wu, and Wandong Zhang.
\newblock Review of accident detection methods using dashcam videos for autonomous driving vehicles.
\newblock {\em IEEE Transactions on Intelligent Transportation Systems}, 2024.

\bibitem{shaikh2024evaluating}
Junaid Shaikh and Nils Lubbe.
\newblock Evaluating forward collision warning and autonomous emergency braking systems in india using dashboard cameras.
\newblock Technical report, SAE Technical Paper, 2024.

\bibitem{sharma2024two}
Bhuvanesh~Kumar Sharma, Aman Sharma, Sanjay~Kumar Sharma, Yogesh Mahajan, and Sneha Rajput.
\newblock Do the two-wheeler safety harnesses effective in rider’s safety- analysis of attitude and switching intention.
\newblock {\em Case Studies on Transport Policy}, 15:101146, 2024.

\bibitem{crnn}
Baoguang Shi, Xiang Bai, and Cong Yao.
\newblock An end-to-end trainable neural network for image-based sequence recognition and its application to scene text recognition.
\newblock {\em IEEE Transactions on Pattern Analysis and Machine Intelligence}, 39(11):2298--2304, 2016.

\bibitem{soltanikazemi2023real}
Elham Soltanikazemi, Ashwin Dhakal, Bijaya~Kumar Hatuwal, Imad~Eddine Toubal, Armstrong Aboah, and Kannappan Palaniappan.
\newblock Real-time helmet violation detection in ai city challenge 2023 with genetic algorithm-enhanced yolov5.
\newblock In {\em 2023 IEEE Applied Imagery Pattern Recognition Workshop (AIPR)}, pages i--x. IEEE, 2023.

\bibitem{sugiarto2021helmet}
Richard Sugiarto, Evan~Kusuma Susanto, and Yosi Kristian.
\newblock Helmet usage detection on motorcyclist using deep residual learning.
\newblock In {\em 2021 3rd East Indonesia Conference on Computer and Information Technology (EIConCIT)}, pages 194--198. IEEE, 2021.

\bibitem{thombre2022automated}
Supriya~S Thombre, Sampada~S Wazalwar, Rakshanda Tarekar, Sakshi Ramgirwar, Shivani Zulkanthiwar, and Shubhra Dubey.
\newblock Automated traffic rule violation detection with e-challan generation for smart societies.
\newblock In {\em Decision Analytics for Sustainable Development in Smart Society 5.0: Issues, Challenges and Opportunities}, pages 65--81. Springer, 2022.

\bibitem{tran2023robust}
Duong Nguyen-Ngoc Tran, Long~Hoang Pham, Hyung-Joon Jeon, Huy-Hung Nguyen, Hyung-Min Jeon, Tai Huu-Phuong Tran, and Jae~Wook Jeon.
\newblock Robust automatic motorcycle helmet violation detection for an intelligent transportation system.
\newblock In {\em Proceedings of the IEEE/CVF Conference on Computer Vision and Pattern Recognition}, pages 5341--5349, 2023.

\bibitem{tsai2023video}
Chun-Ming Tsai, Jun-Wei Hsieh, Ming-Ching Chang, Guan-Lin He, Ping-Yang Chen, Wei-Tsung Chang, and Yi-Kuan Hsieh.
\newblock Video analytics for detecting motorcyclist helmet rule violations.
\newblock In {\em Proceedings of the IEEE/CVF Conference on Computer Vision and Pattern Recognition}, pages 5365--5373, 2023.

\bibitem{van2024motorcyclist}
Thien Van~Luong, Huu Si~Phuc Nguyen, Duy~Khanh Dinh, Viet~Hung Duong, Duy Hong~Sam Vo, Huan Vu, Minh~Tuan Hoang, and Tien~Cuong Nguyen.
\newblock Motorcyclist helmet violation detection framework by leveraging robust ensemble and augmentation methods.
\newblock In {\em Proceedings of the IEEE/CVF Conference on Computer Vision and Pattern Recognition}, pages 7027--7036, 2024.

\bibitem{vo2024robust}
Hao Vo, Sieu Tran, Duc~Minh Nguyen, Thua Nguyen, Tien Do, Duy-Dinh Le, and Thanh~Duc Ngo.
\newblock Robust motorcycle helmet detection in real-world scenarios: Using co-detr and minority class enhancement.
\newblock In {\em Proceedings of the IEEE/CVF Conference on Computer Vision and Pattern Recognition}, pages 7163--7171, 2024.

\bibitem{wang2023prb}
Bor-Shiun Wang, Ping-Yang Chen, Yi-Kuan Hsieh, Jun-Wei Hsieh, Ming-Ching Chang, JiaXin He, Shin-You Teng, HaoYuan Yue, and Yu-Chee Tseng.
\newblock Prb-fpn+: Video analytics for enforcing motorcycle helmet laws.
\newblock In {\em Proceedings of the IEEE/CVF Conference on Computer Vision and Pattern Recognition}, pages 5476--5484, 2023.

\bibitem{wojke2017simple}
Nicolai Wojke, Alex Bewley, and Dietrich Paulus.
\newblock Simple online and realtime tracking with a deep association metric.
\newblock In {\em 2017 IEEE international conference on image processing (ICIP)}, pages 3645--3649. IEEE, 2017.

\bibitem{hybrid}
Mingzhan Yang, Guangxin Han, Bin Yan, Wenhua Zhang, Jinqing Qi, Huchuan Lu, and Dong Wang.
\newblock Hybrid-sort: Weak cues matter for online multi-object tracking.
\newblock In {\em Proceedings of the AAAI Conference on Artificial Intelligence}, volume~38, pages 6504--6512, 2024.

\bibitem{zekany2019classifying}
Stephen~A Zekany, Ronald~G Dreslinski, and Thomas~F Wenisch.
\newblock Classifying ego-vehicle road maneuvers from dashcam video.
\newblock In {\em 2019 IEEE Intelligent Transportation Systems Conference (ITSC)}, pages 1204--1210. IEEE, 2019.

\bibitem{zhang2024coarse}
Hongpu Zhang, Zhe Cui, and Fei Su.
\newblock A coarse-to-fine two-stage helmet detection method for motorcyclists.
\newblock In {\em Proceedings of the IEEE/CVF Conference on Computer Vision and Pattern Recognition}, pages 7066--7074, 2024.

\bibitem{bytetrack}
Yifu Zhang, Peize Sun, Yi Jiang, Dongdong Yu, Fucheng Weng, Zehuan Yuan, Ping Luo, Wenyu Liu, and Xinggang Wang.
\newblock Bytetrack: Multi-object tracking by associating every detection box.
\newblock In {\em European conference on computer vision}, pages 1--21. Springer, 2022.

\bibitem{ciou}
Zhaohui Zheng, Ping Wang, Dongwei Ren, Wei Liu, Rongguang Ye, Qinghua Hu, and Wangmeng Zuo.
\newblock Enhancing geometric factors in model learning and inference for object detection and instance segmentation.
\newblock {\em IEEE transactions on cybernetics}, 52(8):8574--8586, 2021.

\bibitem{zong2023detrs}
Zhuofan Zong, Guanglu Song, and Yu Liu.
\newblock Detrs with collaborative hybrid assignments training.
\newblock In {\em Proceedings of the IEEE/CVF international conference on computer vision}, pages 6748--6758, 2023.

\end{thebibliography}

\clearpage
\renewcommand{\thesection}{\arabic{section}}
{\Large \textbf{Supplementary Material}}
\section{RideSafe-400 Dataset}
\label{sec:dataset_suppl}

\begin{figure*}[t]
\begin{center}
\includegraphics[width=0.9\linewidth]{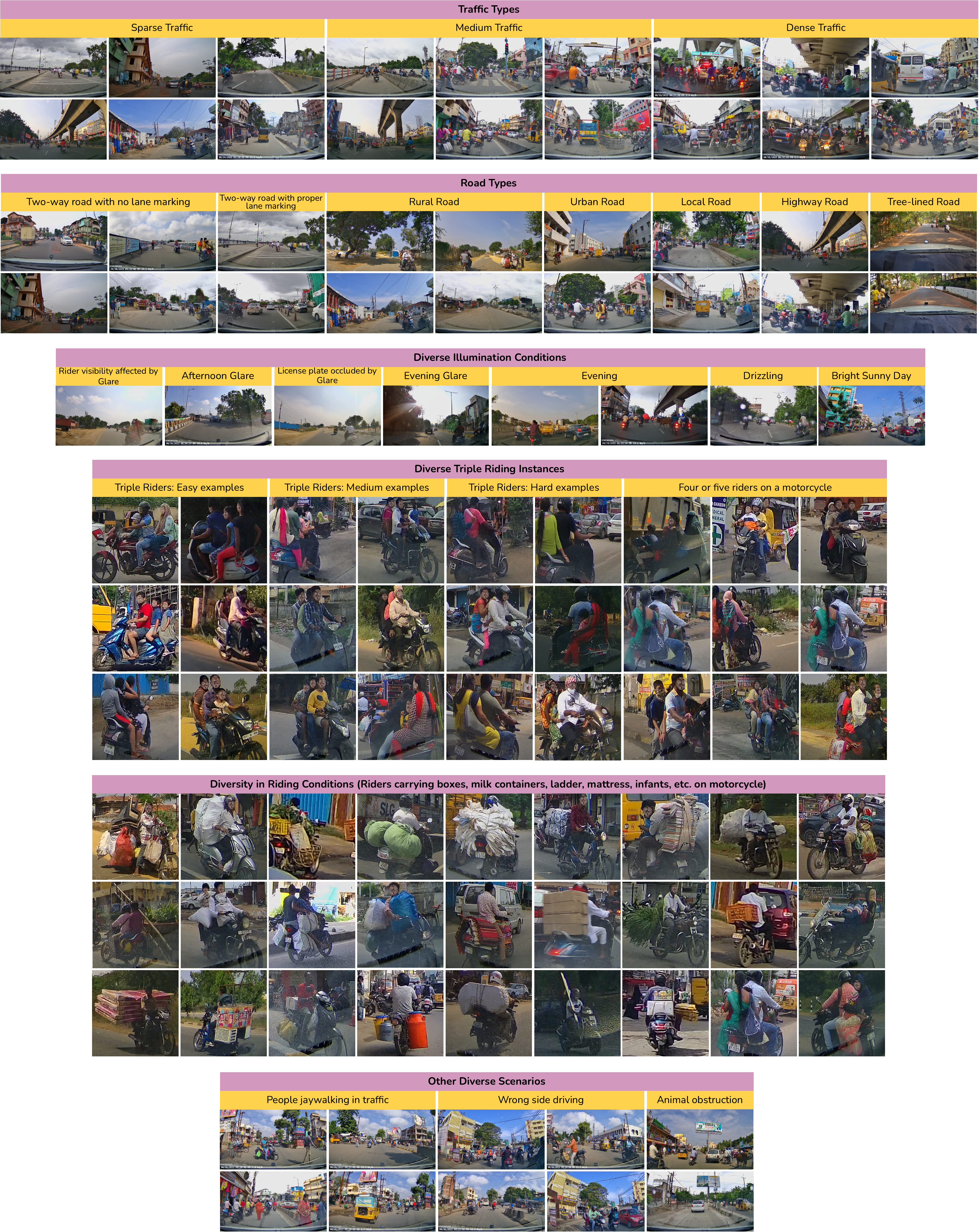}
\caption{Diverse scenarios considered for annotation.}
\label{fig:challenging_scenarios_suppl}
\end{center}
\end{figure*}

Our dataset, RideSafe-400, comprises 400 driving videos specifically curated for identifying triple riding and helmet rule violators on the road. The traffic scenarios are captured using a DDPAI X2S Pro dashcam, with a resolution of $1920 \times 1080$ pixels and a frame rate of 25 fps. Each video spans between 60 to 72 seconds, culminating in approximately 600K frames. These videos contain multiple instances of triple riding or helmet rule violations, captured in a diverse range of traffic conditions, including varying traffic types, street types, illumination, weather, and other challenging scenarios. A more elaborate diagram is illustrated in \cref{fig:challenging_scenarios_suppl}.

\begin{figure}[t]
  \centering  
   \includegraphics[width=\linewidth]{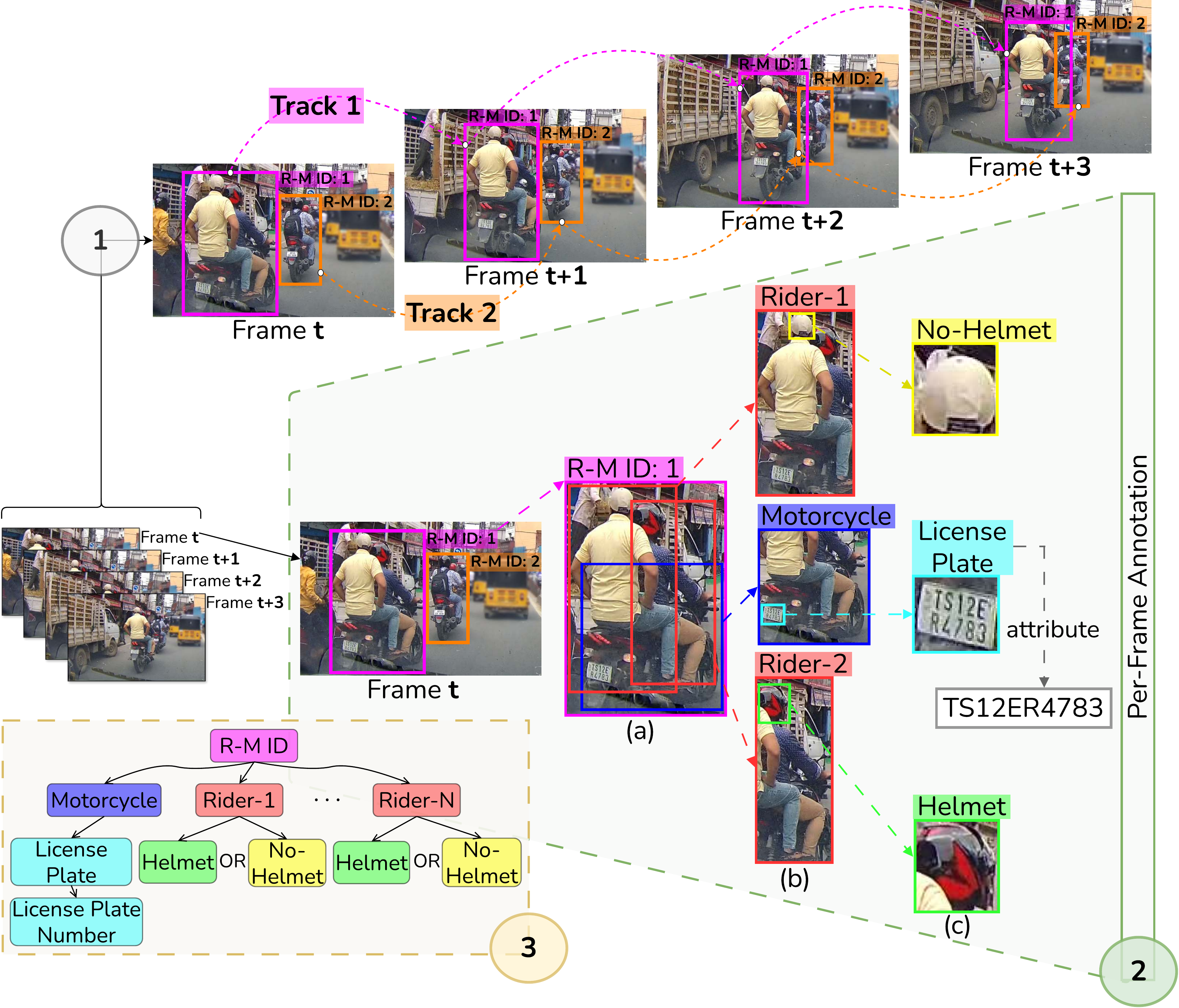}
   \caption{The annotation schema for our RideSafe-400 dataset. Refer to \cref{sec:dataset_suppl} for details.}
   \label{fig:annotation_classes_suppl}
\end{figure}

Annotations in RideSafe-400 cover six object classes: `R-M instance', `rider', `motorcycle', `helmet', `no-helmet', and `license plate' (see \cref{fig:annotation_classes_suppl}-\circled{1}). For each video frame, riders are associated with their respective motorcycles (termed an R-M instance) using rectangular bounding boxes. Each R-M instance is further annotated with the classes rider and motorcycle (see \cref{fig:annotation_classes_suppl}-\circled{2} (a)). For each rider, the presence or absence of a helmet is annotated (\cref{fig:annotation_classes_suppl}-\circled{2} (b)). Each motorcycle object is assigned an attribute `license plate' or `no license plate', depending on the visibility of the license plate, and the corresponding license plate bounding box is annotated along with its plate number (\cref{fig:annotation_classes_suppl}-\circled{2} (c)). To uniquely identify an R-M instance across video frames, a \textit{track id} is provided (see \cref{fig:annotation_classes_suppl}-\circled{1}). Additionally, each R-M instance and its corresponding objects (rider, motorcycle, helmet, no-helmet, license plate) are assigned a common \textit{group id} attribute (see \cref{fig:annotation_classes_suppl}-\circled{3}).

Our dataset includes 354K R-M annotations and 23K R-M tracks, 356K motorcycle annotations and 24K motorcycle tracks, 311K rider annotations and 19K rider tracks, 194K helmet annotations and 10K helmet tracks, 149K no-helmet annotations and 8K no-helmet tracks. Additionally, there are 27K annotated license plates, each labelled with its corresponding license plate number.

\subsection{Dataset Creation Details \& Annotation Procedure}
For annotating the dataset, we utilize the open-source CVAT software. Each video is annotated with six object classes: `R-M instance', `rider', `motorcycle', `helmet', `no-helmet', and `license plate'. Rectangular bounding boxes are used to annotate each class.

\begin{figure}[t]
  \centering
   \includegraphics[width=\linewidth]{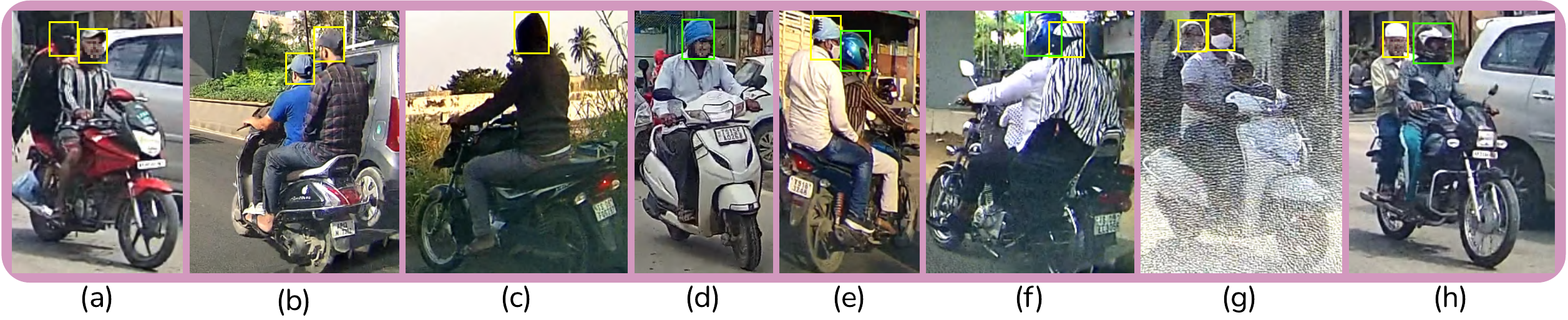}
   \caption{\textbf{Examples of \textit{Helmet} and \textit{No-Helmet} classes}: Annotators were instructed to label a human head as a \textit{helmet} if the individual is wearing a hard plastic protective headgear or a turban, as shown in (d), (e), (f) and (h). All other head coverings, including uncovered heads (g), caps (a) and (b), skull caps (h), hooded garments (c), enveloping garment (a), (f) and (g), and similar items, were annotated as \textit{no-helmet}.}
   \label{fig:hnh_samples}
\end{figure}

\begin{figure}[!t]
  \centering  
   \includegraphics[width=\linewidth]{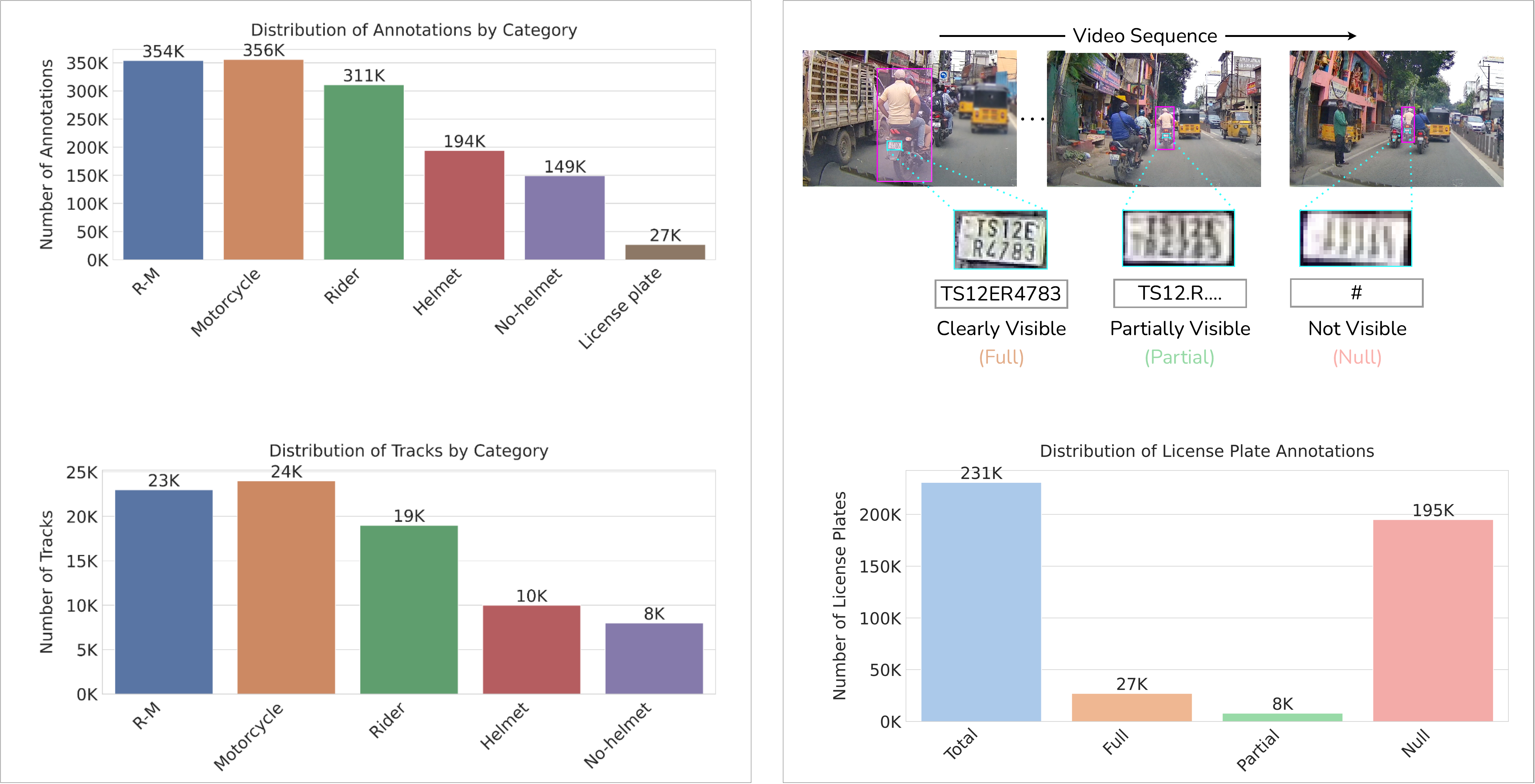}
   \caption{\textbf{Top-Left}: Frame-level annotations for each object category. \textbf{Bottom-Left}: Track-level annotations for each object category. \textbf{Top-Right}: License plate number (LPN) attribute labelling format. \textbf{Bottom-Right}: License plate annotation distribution.}
   \label{fig:distribution_classes_and_lp}
\end{figure}

Annotators were instructed to annotate objects only if they could confidently identify the category. For instance, the distinction between helmet and no-helmet classes becomes ambiguous for distant riders, as those objects appear extremely small and indistinguishable. In such cases, annotators were asked not to annotate those objects. This process is applied to all object classes. A few samples of `helmet' and `no-helmet' are illustrated in \cref{fig:hnh_samples}. 
Additionally, we observed that our dataset has instances of riders wearing a turban. Annotators are asked to label a turban as a `helmet' class to prevent flagging such a case as a helmet rule violation, as shown in \cref{fig:hnh_samples} (d).
Moreover, annotations of small objects below a certain area and aspect ratio threshold are removed. For each class, this threshold is defined based on the average size and aspect ratio of instances of that class in the dataset. A separate set of annotators reviews the annotations to mitigate the potential for cognitive bias and ensure consistency. \cref{fig:distribution_classes_and_lp} shows the distribution of the annotated instances for each object class. Furthermore, our dataset features object-level tracks for each category. \cref{fig:distribution_classes_and_lp} illustrates the distribution of annotated tracks for each class.

For the license plate recognition task, we observed that the visibility of characters varies with distance from the dashcam. Specifically, when a license plate is closer to the dashcam, all characters are typically legible. However, as the distance increases, certain characters become indistinguishable, and beyond a certain threshold, no characters are discernible. To account for this variation in visibility, we introduce a unique annotation schema for the license plate object class, which involves assigning an attribute to the license plate bounding box in each frame, referred to as the \textit{license plate number} (LPN) (\cref{fig:annotation_classes_suppl}-\circled{2} (c)).
The LPN attribute captures the characters visible on the annotated license plate. Depending on the visibility and clarity of these characters, license plates are categorized into three types: (1) Full license plates, where all characters are clearly legible and fully annotated; (2) Partial license plates, where some characters are ambiguous, represented by a placeholder dot symbol ($\cdot$) for each unclear character; and (3) Null license plates, where no characters are visible, and the LPN attribute is assigned a hash symbol (\#). A comprehensive summary of the license plate annotations is presented in \cref{fig:distribution_classes_and_lp}.

We observed that during the assignment of values to the LPN attribute, annotators may infer characters from the adjacent frames in the video sequence, potentially using information from prior or subsequent frames to label characters that are not fully visible. To mitigate this and ensure that characters are labeled only when the annotators are confident, we implemented a two-pass annotation strategy. In the first pass, annotators sequentially review each video frame and label the LPN for every license plate detected. Once all frames have been annotated, in the second pass, we randomly shuffle the frames across the entire video dataset and redistribute them among annotators. Each annotator is then required to re-label the LPN attribute for the shuffled frames. This two-pass process ensures a higher degree of annotation accuracy by reducing potential biases introduced by frame context.

If a character is consistently labeled the same in both passes, the label is accepted. Otherwise, if discrepancies arise, the corresponding character is marked with a placeholder dot symbol ($\cdot$). The shuffling mechanism ensures that annotators do not rely on contextual information from surrounding frames, leading to more accurate and unbiased annotations. The final labeling format is illustrated in \cref{fig:distribution_classes_and_lp}.

\section{Methodology}
\label{sec:methodology_suppl}

\subsection{Segmentation and Cross-Association (SAC)}
\label{sec:SAC_suppl}

\begin{figure*}
\begin{center}
\includegraphics[width=\linewidth]{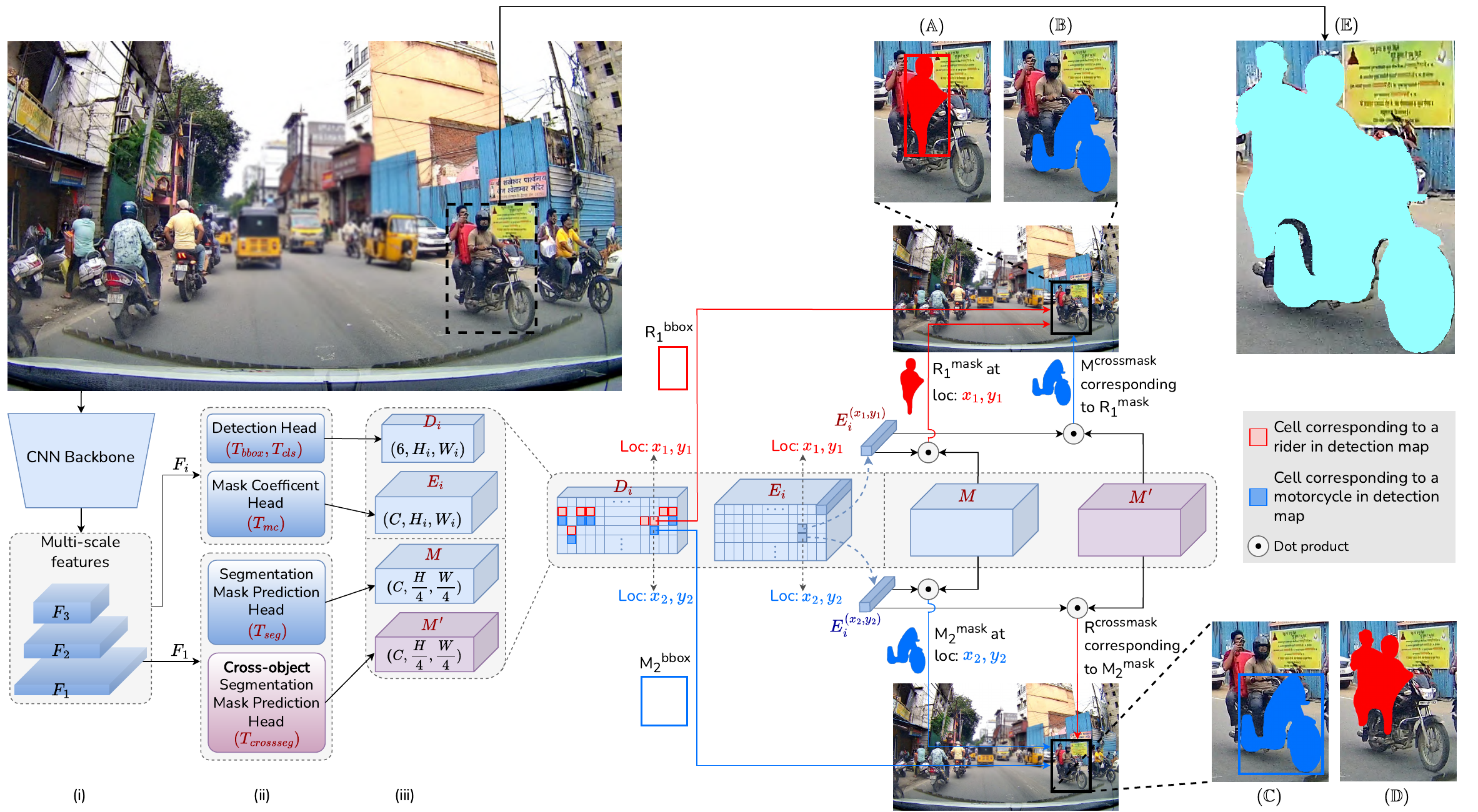}
   \caption{\textbf{Segmentation and Association (SAC):} The input frame (top-left) is processed by the SAC module to predict rider locations ($\mathbb{A}$), motorcycle locations ($\mathbb{C}$), and cross-object segmentation masks ($\mathbb{B,D}$). ($\mathbb{E}$) visualizes the R-M instance. In the architecture diagram, the blue blocks are from YOLO-v8~\cite{Jocher_Ultralytics_YOLO_2023}. The purple blocks are our novel addition which enable association. Refer to Sec.~\ref{sec:SAC_suppl} for details.}
\label{fig:sac_suppl}
\end{center}
\end{figure*}
To achieve precise association between two object classes (rider and motorcycle), along with detecting and segmenting them, we propose a framework that jointly learns both tasks. The features learned for instance segmentation provide valuable information for association, and similarly, the features used for association contribute to improving segmentation. We aim to design an architecture where each class vote for the presence of the other.

In standard convolutional architectures for object detection, the image is divided into a grid, with features from each grid cell used to classify objects, predict bounding boxes, and generate segmentation masks. In our approach, we extend this idea by having the feature vector of a specific grid cell vote for the presence of the other class in the image. We formulate the association between grid cells and the other class as an additional segmentation task. This enables the feature vector to predict not only the segmentation mask of its own class (target object) but also to determine the segmentation mask of the associated class.

\subsubsection{Backbone}
The model comprises a total of 74,017,478 parameters, with the backbone comprising 21 layers, as adopted from YOLOv8 \cite{Jocher_Ultralytics_YOLO_2023}. The backbone architecture primarily consists of convolutional layers responsible for downsampling the input image, followed by upsampling layers. This design ensures that the resulting feature maps retain both high-level and low-level spatial information.

\subsubsection{Heads}
The input to the model is an image \( I \in \mathbb{R}^{3 \times H \times W} \). Through the backbone, the model extracts feature maps from three different layers, denoted as \( F_1 \), \( F_2 \), and \( F_3 \) (corresponding to layers 15, 18, and 21, respectively). The shapes of these features are \( F_1 \in \mathbb{R}^{320 \times \frac{H}{8} \times \frac{W}{8}} \), \( F_2 \in \mathbb{R}^{640 \times \frac{H}{16} \times \frac{W}{16}} \) and \( F_3 \in \mathbb{R}^{640 \times \frac{H}{32} \times \frac{W}{32}} \).

Let each feature map at the $i^{th}$ scale be denoted as $F_{i}$, with shape $(C_i, H_i, W_i)$. To get the bounding box prediction map $B_i$ of shape $(4, H_i, W_i)$ and the class score prediction map $K_i$ of shape $(2, H_i, W_i)$  corresponding to $F_i$, these feature maps are passed through transformation functions $T_{bbox}^{i}$ and $T_{cls}^{i}$. Both transformations are implemented using a series of 3-layer convolutional networks.

\begin{equation}
  B_{i} = T_{bbox}^{i}(F_i) \; \forall \; i \in \{1, 2, 3\}
  \label{eq:supl-one}
\end{equation}

\begin{equation}
    K_{i} = T_{cls}^{i}(F_i) \; \forall \; i \in \{1, 2, 3\}
  \label{eq:supl-two}
\end{equation}

For segmentation and cross-object segmentation, there are three branches. 
The head takes $F_i$ as the input and produces a joint mask coefficient embedding map of shape $(C, H_i, W_i)$ (where C=32 for all experiments).

\begin{equation}
  E_i = T_{mc}^{i}(F_i) \; \forall \; i \in \{1, 2, 3\}
  \label{eq:supl-three}
\end{equation}

The segmentation head $T_{seg}$ and the cross-object segmentation head $T_{crossseg}$ takes $F_1$ as the input and produces an embedding map of shape $(C, \frac{H}{4}, \frac{W}{4})$.

\begin{equation}
  M = T_{seg}(F_1)
  \label{eq:supl-four}
\end{equation}

\begin{equation}
  M' = T_{crossseg}(F_1)
  \label{eq:supl-five}
\end{equation}

Both $T_{seg}$ and $T_{crossseg}$ are implemented through a series of 3-layer convolutional network with an upsampling layer which upsamples the feature map dimensions by a factor of 2.

To get the segmentation mask corresponding to a grid cell at location (x, y) for the $i^{th}$ feature map, we take the dot product between $M$ and $E^{xy}_i$

\begin{equation}
  S_{i}^{xy} = M\cdot E^{xy}_i
  \label{eq:supl-six}
\end{equation}

\begin{equation}
  A_{i}^{xy} = M' \cdot E^{xy}_i
  \label{eq:supl-seven}
\end{equation}

During inference, Non-Maximum Suppression (NMS) is applied to obtain spatial locations in $F_i$ where the class scores exceed a certain threshold. Segmentation and cross-object segmentation masks are then retrieved for these spatial locations, resulting in a set of masks and cross-object masks for both riders and motorcycles. The resulting set of bounding boxes $B^k$, segmentation masks $S^k$, cross-object segmentation masks $A^k$ can be represented as $[(B_r^1, S_r^1, A_r^1), \ldots (B_r^N, S_r^N, A_r^N)]$ for riders and 
$[(B_m^1, S_m^1, A_m^1) \ldots (B_m^M, S_m^M, A_m^M)]$ for motorcycles. The association score $a_{k, l}$ between the $k$-th rider and $l$-th motorcycle detections is computed as

\begin{equation}
\makebox[0.8\columnwidth]{%
    \resizebox{0.9\columnwidth}{!}{$
    a_{k, l} = \frac{1}{2}( \text{IoU}(A_r^k, S_m^l) + \text{IoU}(S_r^k, A_m^l \cdot \text{mask}(B_r^k)))
    $}
}
\label{eq:supl-eight}
\end{equation}

where mask$(B)$ is a function that takes in a bounding box and returns a binary mask with values inside the box set to 1 and outside set to 0. The first term in the equation \cref{eq:supl-eight} represents the IoU between cross-object segmentation mask of the rider and the segmentation mask of the motorcycle. The second term accounts for the IoU between the rider's segmentation mask and the motorcycle's cross-object segmentation mask. Since the cross-object mask of the motorcycle could have more than one riders associated with it, the IoU has to be computed only within the region where the bounding box of the rider is. That is why the cross-object mask of the motorcycle is first restricted to the region of the rider and then the IoU is calculated between this and the rider mask. 
In ideal associations, cross-object masks should perfectly overlap with the segmentation masks of their corresponding associated objects. The association score $a_{k,l}$ ranges from 0 and 1, where 0 indicates no association and 1 represents a perfect association.

\subsubsection{Training}

The bounding box and class predictions are matched to the ground truth bounding boxes and classes using the Task-Alignment Learning (TAL) approach~\cite{feng2021tood}, yielding to a total of $L$ foreground assignments. The matched anchor boxes are evaluated against the ground truth boxes using the Complete IoU (CIOU) loss, $L_{CIOU}$, and the Distribution Focal Loss (DFL), $L_{DFL}$. For the classification task, Binary Cross-Entropy (BCE) loss, $L_{cls}$, is used.

Let $N$ represent the number of riders and $M$ represent the number of motorcycles in the current frame. The ground truth rider masks are denoted by $R_k$ and the motorcycle masks by $M_k$, where $R_k, M_k \in \{0, 1\}^{H \times W}$.
Each rider is assigned an association ID, $r_k$, and similarly, each motorcycle is assigned $m_k$. In any given rider-motorcycle (R-M) instance, the association IDs of all riders match the association ID of the corresponding motorcycle. If no rider corresponds to a motorcycle, $m_k$ is set to -1, and similarly, if no motorcycle corresponds to a rider (due to occlusion), then $r_k$ is set as -1.

Both the segmentation and the cross-object segmentation tasks employ a per-pixel BCE loss. If the $l^{th}$ cell at position $(x,y)$ in the $i^{th}$ grid is assigned to the $k^{th}$ rider ground truth object, then the target mask, $T_l$ for the corresponding associated motorcycle object is the one where $m_k = r_k$. If $r_k = -1$, then $T_l = \textbf{0}$.
If instead, the $l^{th}$ cell is assigned to the $k^{th}$ motorcycle ground truth object, the corresponding target cross-object segmentation mask, $T_l$ is defined as the combined mask of the set

\begin{equation}
 T_l = \{R_j | r_j = m_k \; \forall \; j \in \{1 ... N\}\}
  \label{eq:supl-nine}
\end{equation}

$T_l$ is compared with the predicted cross-object segmentation mask $P_l = A_{i}^{xy}$ for that cell.
The overall cross-object segmentation loss is given by

\begin{equation}
 L_{crossseg} = \sum_{l=1}^{L} L_{BCE}(P_l, T_l)
  \label{eq:supl-ten}
\end{equation}

The overall loss function is
L = $\lambda_1 L_{CIOU} + \lambda_2  L_{DFL} + \lambda_3 L_{cls} + \lambda_4 L_{seg} + \lambda_5 L_{crossseg}$
where $\lambda_1, \lambda_2, \lambda_3, \lambda_4, \lambda_5$ are hyperparameters for balancing multi-task learning objectives.

\subsection{Rider-Motorcycle Association-based Tracking}
\label{sec:rmtracker_supp}

Multiple object tracking (MOT) refers to the task of tracking multiple objects, potentially of different classes, across a sequence of frames. One of the most prominent, efficient, and robust tracking families is the SORT family. Tracking within this framework involves assigning existing tracks to new detections in the current frame. The tracker models the probability distribution $p(x_{T+1} | x_{1:T})$, where $x_i$ represents the state of the object at the $i^{th}$ timestep, where the state of the object includes both appearance and kinematic states. Instead of directly parametrizing a probablity function, it is often more convinient to define a score function or energy function, $s(x_{T+1}, x_{1:T} ; \theta)$, which is not constrained to sum to 1 and can take both positive and negative values. It represents and quantifies the likelihood of associating a track with a detection. Both the formulations are equivalent and are related by the expression

\begin{equation}
    p(x_{T+1} | x_{1:T}) = \frac{e^{s(x_{T+1}, x_{1:T} ; \theta)}}{\sum_{x_{T+1}} e^{s(x_{T+1}, x_{1:T} ; \theta) }}
    \label{eq:supl-eleven}
\end{equation}

where the summation can be replaced by an integral if the score function is continuous.
The objective of the tracker is to determine an optimal combination of track-to-detection assignments that maximizes the overall probability of association. This overall probability is defined as the product of the probabilities for each corresponding track-detection pair. The problem can be formulated as a bipartite matching between the current set of tracks and the available detections by constructing a score matrix that quantifies the likelihood of associating the $i^{th}$ track with the $j^{th}$ detection using a predefined score function. Each possible assignment between tracks and detections is represented by a binary variable $a_{ij} \in {0, 1}$, where $s_{ij}$ denotes the score of assigning the $i^{th}$ track with the $j^{th}$ detection. The goal is to maximize

\begin{equation}
    max \sum_{i=1}^{i=N}\sum_{j=1}^{j=M} a_{ij}\cdot{s_{ij}}
    \label{eq:supl-twelve}
\end{equation}

subject to the constraints

\begin{equation}
    \sum_{i=1}^{i=N} a_{ij} \leqslant 1 \; \forall \; j \in \{1 \ldots M\}
    \label{eq:supl-thirteen}
\end{equation}

\begin{equation}
    \sum_{j=1}^{j=M} a_{ij} \leqslant 1 \; \forall \; i \in \{1 \ldots N\}
    \label{eq:supl-fourteen}
\end{equation}

These constraints ensure that each detection is assigned to at most one track and vice versa. This formulation corresponds to the well-known linear sum assignment problem, which can be solved efficiently. This problem is independently addressed for each class. The score $s_{ij}$ is a function of a motion model, typically a Kalman filter, and an appearance model, such as an object re-identification model.
Equivalently the above objective and constraint can be seen as maximizing the total probability of association of the existing tracks to the detections in the current frame.

For the downstream task of triple riding detection, instead of tracking each class independently, we aim to jointly track riders and motorcycles while simultaneously capturing their evolving associations. This joint approach offers two advantages. It improves the tracking accuracy for each object class, and establishes robust, persistent associations between rider and motorcycle instances across time.

For a general two-class tracking problem, the objective of the tracker extends to modeling the joint probability distribution, $p(x_{t+1}, y_{t+1} | x_{1:t}, y_{1:t})$. where $x$ and $y$ represents the state of the two classes. Instead of making the class independence assumption

\begin{equation}
    p(x_{t+1}, y_{t+1} | x_{1:t}, y_{1:t}) = p(x_{t+1} | x_{1:t}) \cdot p(y_{t+1} | y_{1:t})
    \label{eq:supl-fifteen}
\end{equation}

we use cross-class association scores to model the dependencies between $p(y_t | x_t)$ and $p(x_t | y_t)$. The updated model is expressed as:

\begin{equation}\label{eq:pgm_model}
\begin{aligned}
    & p(x_{t+1}, y_{t+1} \mid x_{1:t}, y_{1:t}) =  p(x_{t+1} \mid x_{1:t}) \cdot p(y_{t+1} \mid y_{1:t}) \\
    & \quad \cdot \{p(x_{t+1} \mid y_{t+1}) \cdot p(y_{t+1} \mid x_{t+1}) \\ 
    & \quad \cdot p(x_{1:t} \mid y_{1:t}) \cdot p(y_{1:t} \mid x_{1:t})\}
\end{aligned}
\end{equation}

\begin{figure}[t]
  \centering
   \includegraphics[width=0.7\linewidth]{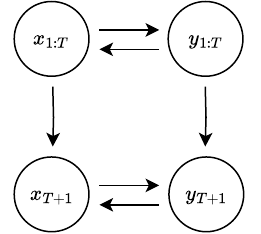}
   \caption{The probabilistic graphical model of the proposed tracker. See \cref{eq:pgm_model}}
   \label{fig:pgm}
\end{figure}

The probabilistic graphical model of the aforementioned formulation is shown in \cref{fig:pgm}. The first two terms correspond to individual class state evolution, and the third term incorporates cross-class associations. Each term in the factorized form above can also be expressed using an equivalent score or energy-based formulation.

Cross-class association scores provide additional insight into the tracking process. Intuitively, if the tracking of one class (\eg motorcycles) is highly confident and its association with another class (\eg riders) is well-established, then tracking the second class becomes more accurate, even when independent tracking of the second class is challenging. Therefore, the overall score for any combination depends not only on the appearance and motion of each class but also on their association with the other object class.

Let the set of rider tracks till time $t$, be represented by

\begin{equation}
    T^r_t = \{r_1, r_2, \dots, r_{N_r^t}\}
\end{equation}

and the set of motorcycle tracks by

\begin{equation}
    T^m_t = \{m_1, m_2, \dots, m_{N_m^t}\}
    \label{eq:supl-sixteen}
\end{equation}

where $N_r^t$ and $N_m^t$ denote the number of rider and motorcycle tracks at time $t$, respectively. The set of rider detections at time $t+1$ is denoted $D^r_{t+1}$, and for motorcycles, $D^m_{t+1}$. The possible assignments for the rider and motorcycle classes are given by

\begin{equation}
    C^r = \{(r_i, d_j) | r_i \in T^r_t \; \& \; d_j \in D^r_{t+1}\}
    \label{eq:supl-seventeen}
\end{equation}

\begin{equation}
    C^m = \{(m_i, d_j) | m_i \in T^m_t \; \& \; d_j \in D^m_{t+1}\}
    \label{eq:supl-eightteen}
\end{equation}

We define a likelihood score based on association alone for each pair of combinations, one from $C^r$ and one from $C^m$. For the $i^{th}$ rider hypothesis in $C^r$ and the $j^{th}$ motorcycle hypothesis in $C^m$, we define a binary variable $e_{ij}$ with an associated score $a_{ij}$, reflecting the probability of association of these two hypotheses

\begin{equation}
    (\{p(x_{t+1} | y_{t+1}) \cdot p(y_{t+1} | x_{t+1}) \cdot p(x_{1:t} | y_{1:t}) \cdot p(y_{1:t} | x_{1:t}\})
    \label{eq:supl-nineteen}
\end{equation}

\begin{figure}[t]
  \centering
   \includegraphics[width=\linewidth]{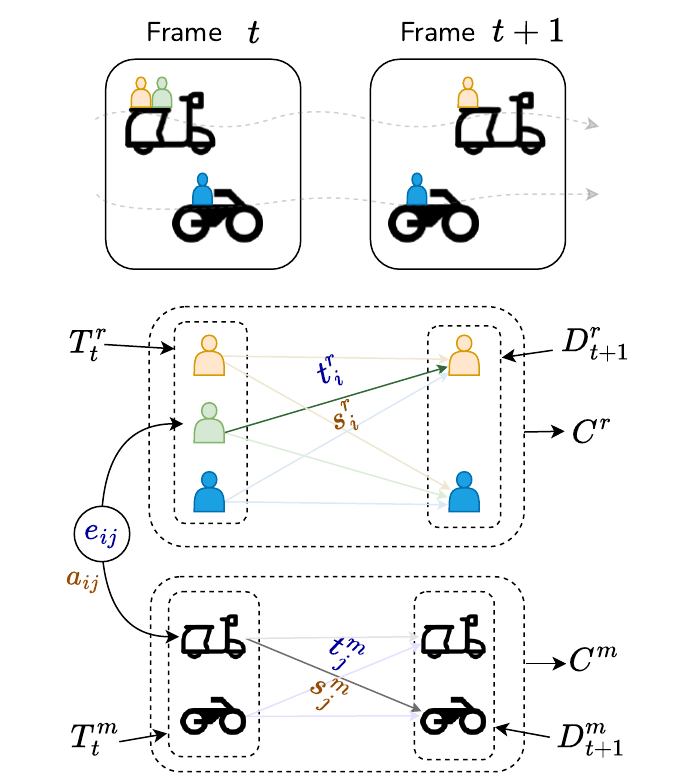}
   \caption{\textbf{R-M Association-based Tracking Module} : Pictorial representation of tracking operation. Notations in blue correspond to the variables, orange correspond to the scores, and the black notations correspond to the sets. Refer to \cref{sec:rmtracker_supp} for details.}
   \label{fig:onecol}
\end{figure}

The value of $e_{ij}$ should be set to 0 if in the final solution, either of $C^r_i$ and $C^m_j$ are not chosen. The value of $e_{ij}$ is constrained such that $e_{ij} = 1$ only when both $C^r_i$ and $C^m_j$ are chosen. This can be represented as

\begin{equation}
    e_{ij}(1 - r_i \cdot m_j) = 0 \; \forall \; i,j
    \label{eq:supl-twenty}
\end{equation}

or equivalently by introducing auxiliary variables

\begin{equation}
    e_{ij}(1 - z_{ij}) = 0
    \label{eq:supl-twentyone}
\end{equation}

\begin{equation}
    z_{ij} = r_i \cdot m_j
    \label{eq:supl-twentytwo}
\end{equation}

Additionally, the constraint that one rider can only be associated with one motorcycle is added

\begin{equation}
    \sum_{i=1}^{i=|C^m|} e_{ij} \leq 1 \; \forall \; j
    \label{eq:supl-twentythree}
\end{equation}

subject to these additional constraints.

In our modified formulation, we are interested in

\footnotesize  

\begin{equation}
\begin{aligned}
(t^{r*}, t^{m*}, e^{*}) =  \underset{\{t^r_i, t^m_j, e_{ij}\} \in \{0, 1\}}{\arg\max} \; &\left( \lambda_1 \sum_{i=1}^{|C_r|} t_{i}^{r} s_i^r + \lambda_2 \sum_{j=1}^{|C_m|} t_{j}^{m} s_j^m \right. \\
&\left. + \lambda_3 \sum_{i=1}^{|C_r|} \sum_{j=1}^{|C_m|} e_{ij} a_{ij} \right)
\end{aligned}
\end{equation}
\normalsize  

subject to the aforementioned constraints. Here, $s_i^r$ is the score for $i^{th}$ rider hypothesis and $s_j^m$ is the score for $j^{th}$ motorcycle hypothesis. The values obtained by the variables, $t_i^r$ and $t_j^m$ after the optimization will reflect the next set of tracks at time $t+1$, and will be associated based on the value of $e_{ij}$.

\section{Training and Implementation Details}

\subsection{License Plate Detection and Recognition}
A YOLOv8-x model~\cite{Jocher_Ultralytics_YOLO_2023} is trained for license plate detection. To improve performance during both training and inference, multi-line license plates of motorcycles are converted into a single-line format. License plate recognition is performed using an OCR model~\cite{crnn}, which is trained on real-world, augmented, and synthetic license plate data. Specifically, we train the CRNN model for 5000 iterations with a batch size of 128, using AdamW optimizer with a learning rate of 0.01.

\subsubsection{Synthetic Plate Generation and Augmentation for Training OCR}
\label{sec:synthlicense}
To enhance the diversity and magnitude of the license plate dataset, license plates from the RiseSafe-400 dataset were augmented to create various perspectives. The augmentations that were performed were perspective transformation (forward, backward, sideways, tilt), increase in contrast, and rotation.

\begin{figure}[t]
\begin{center}
\label{sec:synth_qual}
\includegraphics[width=\linewidth]{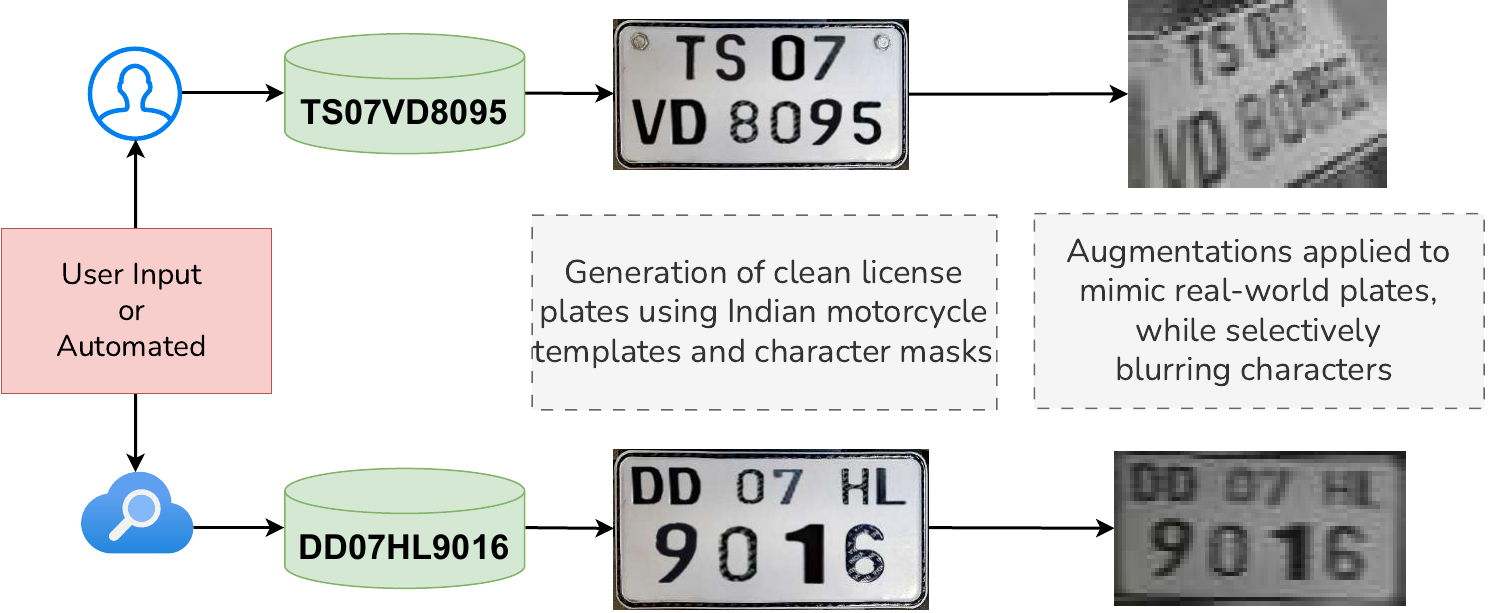}
   \caption{Synthetic license plate dataset generation pipeline. Refer to Sec.~\ref{sec:synthlicense} for details. The user can also give a prompt for the generation of the license plates.}
\label{fig:synth_qual}
\end{center}
\end{figure}

In addition to the above, we developed a pipeline for synthetic license plate generation. This pipeline takes into consideration the numeric patterns of the license plate numbers in the real plates in order to mimic the real-world character distribution. We observe that our dataset has seven types of license plate templates. We first extract the character masks across these seven templates, then to generate a new license plate, we use the extracted character masks and position them on a new plate (See \cref{fig:synth_qual}). 
Once the clean synthetic license plates were generated, a random amount of noise and blur was added in order to resemble real-world plates, along with the same type of augmentations mentioned above. The types of augmentations used were applied for both real-world plates and synthetic plates. 
As seen in the ablations for the license plate OCR model (\cref{tab:lpr_results}), the training dataset that comprised of the real-world, augmented real-world, and synthetic plates gave the best CER on the validation set of our dataset.

\begin{figure*}
\begin{center}
\includegraphics[width=\linewidth]{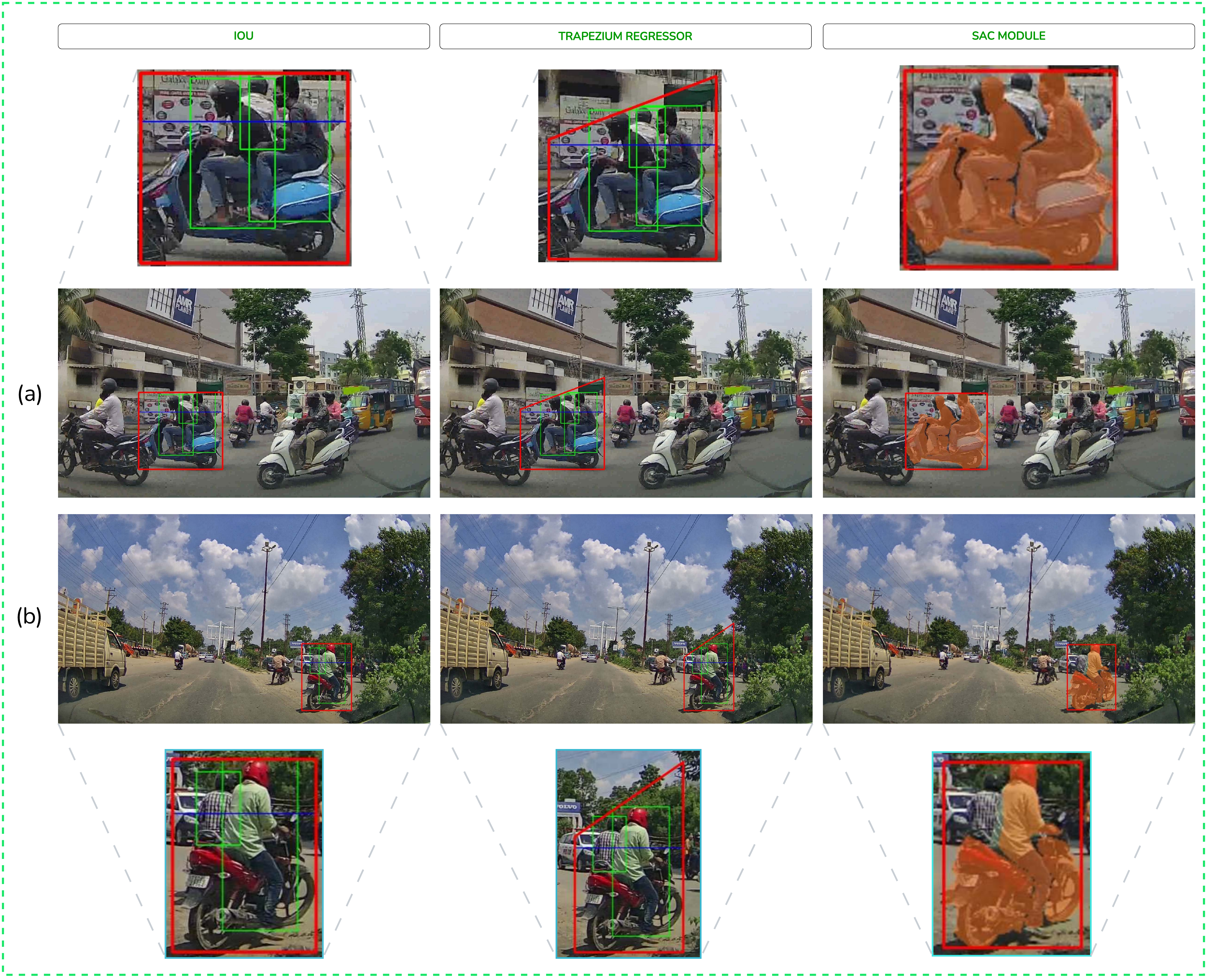}
   \caption{R-M Association Qualitative Analysis.}
\label{fig:assoc_qual}
\end{center}
\end{figure*}

\begin{figure*}
\begin{center}
\includegraphics[width=\linewidth]{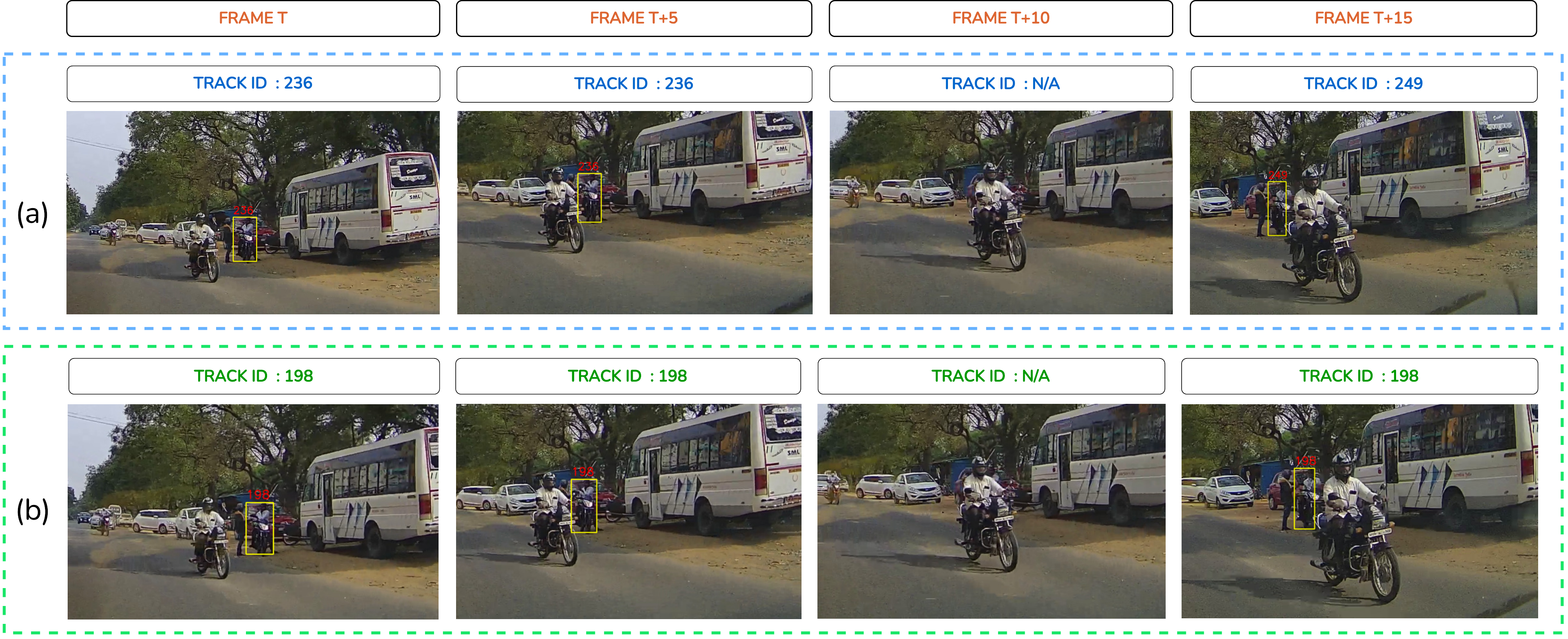}
   \caption{R-M Tracking Qualitative Analysis.}
\label{fig:tracker_qual}
\end{center}
\end{figure*}

\begin{figure*}
\begin{center}
\label{sec:tr_qualitative}
\includegraphics[width=\linewidth]{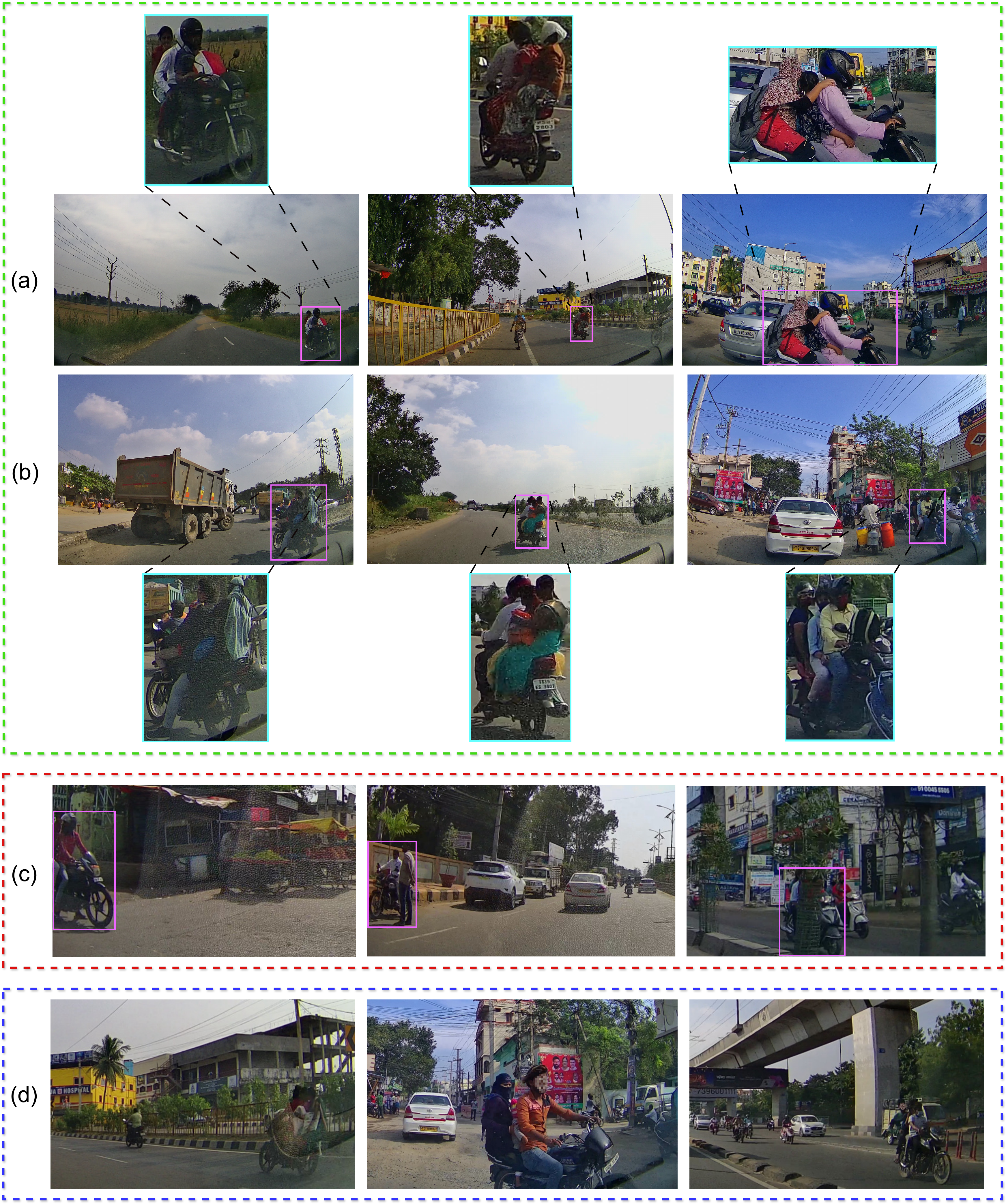}
   \caption{Triple Riding Detection Qualitative Analysis.}
\label{fig:tr_qual}
\end{center}
\end{figure*}

\begin{figure*}
\begin{center}
\includegraphics[width=0.75\linewidth]{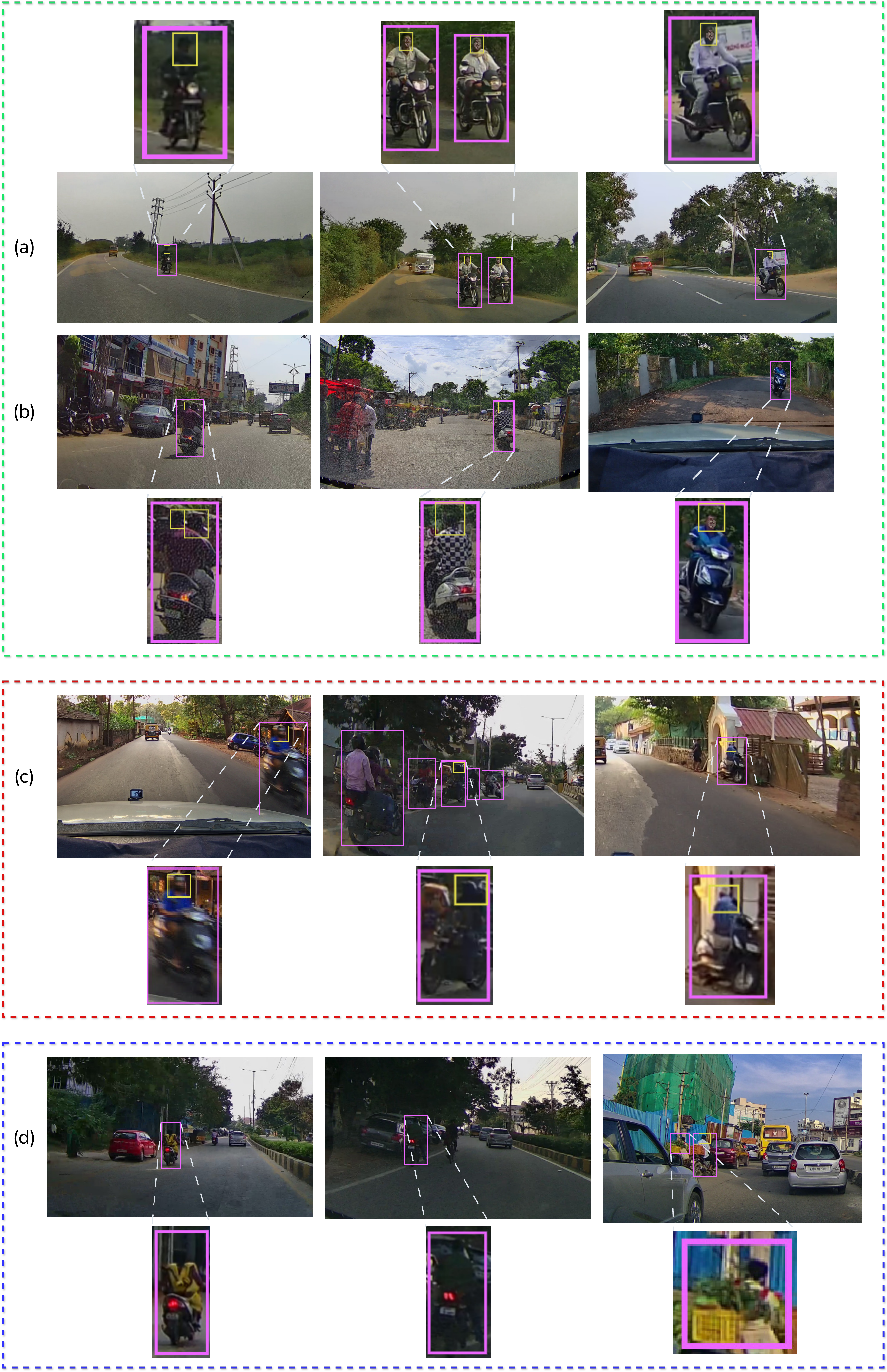}
   \caption{No Helmet Detection Qualitative Analysis.}
\label{fig:hnh_qual}
\end{center}
\end{figure*}

\section{Qualitative Analysis}
\subsection{Segmentation and Cross-Association (SAC)}
In \cref{fig:assoc_qual}, we present a qualitative comparison between the Intersection over Union (IoU) method, the trapezium method \cite{goyal2022detecting}, and the proposed SAC method. The results demonstrate the superiority of SAC, particularly in highly dense traffic scenarios. In the first example (\cref{fig:assoc_qual} (a)), both IoU and the trapezium-based methods incorrectly associate the middle rider as a part of the R-M instance, when clearly the middle rider is not associated with the motorcycle. Both the IoU and the trapezium based methods does not take into account the visual information, and directly operates on the coordinates of the bounding boxes present. In contrast, SAC learns associations directly from pixel-level features, leading to more accurate associations. A similar example is shown in \cref{fig:assoc_qual} (b), where SAC correctly associates the rider and motorcycle.

\subsection{R-M Association-based Tracking}

\cref{fig:tracker_qual} compares between BotSORT~\cite{botsort} and our proposed Cross-AssociationSORT. The BotSORT tracker loses track of the instance when occluded by a passing vehicle. In contrast, Cross-AssociationSORT provides more robust handling of occlusions by jointly tracking both riders and motorcycles as it provides a more confident track-to-detection correspondence.

\subsection{Triple Riding Detection}

In \cref{fig:tr_qual} (a) and (b), we see that the model successfully detects triple riders across various challenging conditions, such as different weather and road types, as well as occlusions.
In \cref{fig:tr_qual} (c), we see some false positives by the system, where the model mistakenly predicts violations. The images have a pattern of heavy occlusion where the instance is obstructed by some other object in the environment.
\cref{fig:tr_qual} highlights missed violations, that the model failed to identify. Such cases arise when the footprint of the occluded rider (often the middle rider) is too small to provide sufficient visual cues for correct identification.

\subsection{Helmet Detection}

In \cref{fig:hnh_qual} (a) and (b), the model correctly identifies helmet violations. In \cref{fig:hnh_qual} (c), we see some examples of false positives. In the image on the left, due to motion blur, a false `no-helmet' violation is detected. In the second false positive, the R-M instance is quite far and the crop is not well illuminated. In the third figure, a pedestrian is incorrectly associated with a motorcycle.
In \cref{fig:hnh_qual} (d), the first 2 false negatives are due to blur and low light conditions, and the the last one is due to the occluded motorcycle hence reducing the context in no helmet detection.

\end{document}